\documentclass[10pt,twocolumn,letterpaper]{article}
\PassOptionsToPackage{dvipsnames}{xcolor}
\usepackage[most]{tcolorbox}
\usepackage{booktabs}
\usepackage{adjustbox}
\usepackage{array}
\usepackage{multirow}
\usepackage{makecell} 
\usepackage{colortbl}
\usepackage{dirtytalk}
\usepackage[ruled,linesnumbered]{algorithm2e}

\DeclareUnicodeCharacter{2206}{$\Delta$} 
\DeclareUnicodeCharacter{2013}{--}       
\DeclareUnicodeCharacter{00A0}{~}     

\usepackage[pagenumbers]{wacv} % To force page numbers, e.g. for an arXiv version

\definecolor{lightblue}{RGB}{16,98,180}
\definecolor{lightpink}{RGB}{243,40,109}
\definecolor{lightgreen}{RGB}{0,220,0}
\AtBeginDocument{%
  \hypersetup{%
    linkcolor=lightpink,
    citecolor=lightgreen,
    urlcolor=lightpink,
    filecolor=lightblue
  }%
}
\definecolor{wacvblue}{rgb}{0.21,0.49,0.74}
\usepackage[pagebackref,breaklinks,colorlinks,allcolors=wacvblue]{hyperref}

\def\wacvPaperID{1010} % *** Enter the WACV Paper ID here
\def\confName{WACV}
\def\confYear{2026}

\title{Knowledge to Sight: Reasoning over Visual Attributes via Knowledge Decomposition for Abnormality Grounding}

\newcommand{\midsize}{\fontsize{10.2pt}{11pt}\selectfont}
\author{
\makebox[\textwidth][c]{%
Jun Li$^{1,2}$ \quad
Che Liu$^{3}$ \quad
Wenjia Bai$^{3}$ \quad
Mingxuan Liu$^{4}$%
} \\
\makebox[\textwidth][c]{%
Rossella Arcucci$^{3}$ \quad
Cosmin I. Bercea$^{1,5*}$ \quad
Julia A. Schnabel$^{1,2,5,6}$\thanks{Shared senior authors.}%
} \\[0.3em]
\makebox[0.4\textwidth][c]{%
\midsize$^{1}$ Technical University of Munich \quad
$^{2}$ Munich Center for Machine Learning \quad
$^{3}$ Imperial College London
} \\
\makebox[\textwidth][c]{%
\midsize$^{4}$ University of Trento \quad
$^{5}$ Helmholtz AI and Helmholtz Munich \quad
$^{6}$ King’s College London
}
}

\begin{document}
\maketitle

\definecolor{OursBlue}{RGB}{16,98,180}
\definecolor{SOTAPink}{RGB}{243,40,109}
\begin{abstract}
In this work, we address the problem of grounding abnormalities in medical images, where the goal is to localize clinical findings based on textual descriptions. While generalist Vision-Language Models (VLMs) excel in natural grounding tasks, they often struggle in the medical domain due to rare, compositional, and domain-specific terms that are poorly aligned with visual patterns. Specialized medical VLMs address this challenge via large-scale domain pretraining, but at the cost of substantial annotation and computational resources.
To overcome these limitations, 
we propose \textbf{Knowledge to Sight (K2Sight)}, a framework that introduces structured semantic supervision by decomposing clinical concepts into interpretable visual attributes, such as shape, density, and anatomical location. These attributes are distilled from domain ontologies and encoded into concise instruction-style prompts, which guide region-text alignment during training. Unlike conventional report-level supervision, our approach explicitly bridges domain knowledge and spatial structure, enabling data-efficient training of compact models.
We train compact models with 0.23B and 2B parameters using only 1.5\% of the data required by state-of-the-art medical VLMs. Despite their small size and limited training data, these models achieve performance on par with or better than 7B+ medical VLMs, with up to 9.82\% improvement in $mAP_{50}$. Code and models: \href{https://lijunrio.github.io/K2Sight/}{\textcolor{SOTAPink}{https://lijunrio.github.io/K2Sight/}}.

\end{abstract}
\begin{figure}[ht]
    \centering
    \includegraphics[width=\linewidth]{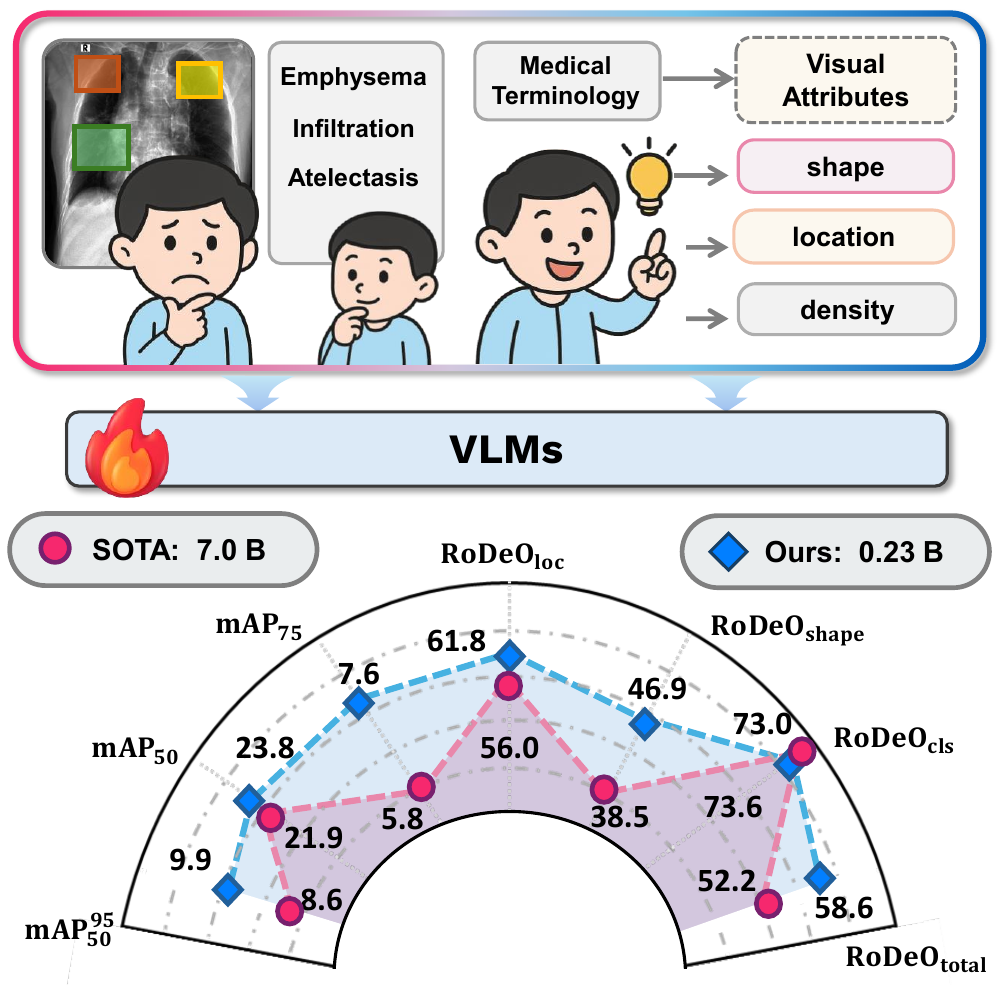}
    \caption{Overview of Knowledge to Sight (K2Sight).
    \textbf{(Top)}: Complex medical terminology is distilled into attribute-based visual instructions, bridging knowledge to sight for abnormality grounding.
   \textbf{(Bottom)}: Our {\color{OursBlue}K2Sight-Lite}, a 0.23B model trained on only 1.5\% of the data, outperforms the 7.0B {\color{SOTAPink}SOTA} medical VLM trained on 1 million samples.
    }
    \label{teaser}
\end{figure}

\section{Introduction}

Vision Language Models (VLMs)~\cite{achiam2023gpt,xiao2024florence,bai2023qwen,lu2024deepseek,zhu2025internvl3} have achieved remarkable success in a variety of visual understanding tasks, such as image captioning~\cite{stefanini2022show}, visual question answering~\cite{lin2023medical}, and visual grounding~\cite{xiao2024towards}. By jointly modeling both visual and textual representations, these models excel at associating image content with natural language descriptions, enabling them to be highly adaptable across diverse domains. Recent advancements have extended VLMs to the medical imaging domain, where models such as RadVLM~\cite{deperrois2025radvlm}, MAIRA-2~\cite{bannur2024maira}, and ChexAgent~\cite{chen2024chexagent} have demonstrated significant potential in tasks like radiology report generation, question answering, and abnormality grounding. These models use large-scale paired X-ray image-text datasets~\cite{johnson2019mimic,wu2021chest,bustos2020padchest,castro2024padchest}, allowing them to perform various medical tasks, thereby providing valuable support to radiologists in clinical diagnosis.

Despite these progress, abnormality grounding remains an underexplored and uniquely challenging task in medical image analysis. Unlike report generation~\cite{li2024ultrasound, tanida2023interactive,li2022self} or question answering~\cite{singhal2025toward,li2025language,zhang2023pmc}, which rely on coarse semantic understanding, abnormality grounding requires precise localization of pathologies described by complex clinical terms. Generalist VLMs often underperform on this task due to their limited exposure to medical concepts and lack of alignment between abstract terminology and nuanced visual features~\cite{LU2025103514}. Specialized medical VLMs alleviate this issue by leveraging extensive medical data and undergoing further supervised finetuning on domain-specific data. However, these models typically involve billions of parameters and large-scale annotations, limiting their usability in data-scarce or resource-constrained settings.

A critical barrier to generalist VLMs in the medical domain lies in the lack of semantically structured supervision. While coarse report descriptions or discrete labels provide high-level semantics, they fail to capture the essential and structural visual primitives such as spatial location, shape, and radiodensity, which radiologists depend on when making localization decisions. We argue that bridging this semantic gap requires decomposing latent clinical concepts into visually grounded representations. These representations can offer a strong inductive bias for grounding tasks.
Instead of relying on monolithic labels, we propose to factorize complex clinical terminology into structured visual attributes and to use these attributes to construct interpretable, instruction-style prompts that guide localization models. This structured supervision enriches visual learning with fine-grained semantics, facilitating data-efficient training and enabling compositional generalization.

In this work, we present K2Sight, a two-stage framework that transforms diagnostic knowledge into visually aligned attribute descriptions, and uses these structured prompts to supervise abnormality grounding. 
K2Sight reformulates long-tail diagnostic terms into interpretable visual attributes, including shape, radiodensity, intensity pattern, and anatomical location. These attributes serve as proxy representations that generalist vision-language models can more effectively learn from. They are automatically derived from standardized radiology ontologies and encoded as concise textual prompts. Compared with visual prototypes, textual prompts are easier to compose, extend, and align across modalities, making them better suited for transfer in low-resource medical settings. This design improves vocabulary coverage, supports compositional generalization, and enables efficient learning from minimal data.

Building on this design, K2Sight applies attribute-based supervision to the grounding process. It decomposes abnormality descriptions into attribute tuples, formats them into structured prompts, and uses these prompts to supervise region-text alignment. Because the attributes reflect the same visual cues that radiologists use in practice, the model can accurately localize abnormalities even with limited supervision. Our experiments demonstrate that knowledge-guided supervision can rival and even surpass scaling-based strategies. With just 0.23 billion parameters and 1.5\% of the training data, our compact K2Sight model outperforms larger medical VLMs with 7 billion parameters—improving $mAP_{50}$ by up to 9.82 percentage points (Fig.~\ref{teaser}).

\begin{itemize}
    \item \textbf{Framework.} 
    We propose a novel framework that reformulates domain knowledge into visually grounded semantic attributes and uses them to supervise grounding, representing a new form of attribute-aligned vision-language supervision.
    This formulation narrows the gap between abstract medical language and spatial image content, improving alignment in model training.

    \item \textbf{Efficiency and Scalability.} We challenge the conventional belief that scaling data and model size is the only path to better performance by demonstrating that our proposed compact K2Sight models (0.23B and 2B parameters), trained on only 1.5\% of the data, can outperform medical VLMs with up to 7 B parameters.

    \item \textbf{Generalization.} Through extensive ablation and transfer experiments, we show that attribute-level supervision substantially enhances abnormality grounding and enables strong zero-shot localization on previously unseen findings. This demonstrates that well-designed supervision can generalize effectively and outperform larger models trained with more resources.

\end{itemize}

\section{Related Work}

\noindent\textbf{General VLMs and the Gap in Abnormality Grounding.} Recent VLMs~\cite{team2023gemini, bai2023qwen, zhu2025internvl3, xiao2024florence} have demonstrated strong generalization across a wide range of visual understanding tasks. By unifying diverse modalities and objectives within a language generation framework, VLMs have introduced a paradigm shift in multimodal learning, enabling elegant and scalable solutions for different tasks like image captioning~\cite{stefanini2022show}, visual question answering~\cite{kuang2025natural}, and open-vocabulary detection~\cite{xiao2024towards}. Recent studies further confirm that generalist VLMs can transfer effectively to the medical domain in a zero-shot setting~\cite{ye2024gmai, arora2025healthbench}, achieving promising results on tasks such as diagnosis and radiology report generation~\cite{lin2023medical, li2024ultrasound}. 

However, their ability to perform abnormality grounding, defined as the spatial localization of pathological findings in medical images, remains limited. 
Although some recent VLMs, such as Qwen2-VL-Instruct~\cite{bai2023qwen}, InternVL~\cite{zhu2025internvl3}, Gemini~\cite{team2023gemini} are capable of abnormality grounding, our evaluations (Sec.~\ref{sec:comparison}) indicate that their zero-shot performance on medical domain is still inadequate. This limitation could be attributed to the absence of domain-specific understanding supervision and the inherent difficulty of aligning complex clinical terminology with subtle and context-dependent visual patterns in radiology. 

\noindent\textbf{Medical-Specific VLMs and the Need for Efficient Grounding.}  
To address the domain shift in medical image understanding, recent work has extended general-purpose VLMs to the medical domain through supervised fine-tuning on large-scale medical datasets~\cite{johnson2019mimic,wu2021chest,demner2016preparing,chambon2024chexpert,nguyen2022vindr}. Earlier models~\cite{moor2023med, alkhaldi2024minigpt, li2023llava} primarily target high-level language tasks, such as report generation~\cite{tanida2023interactive, li2022self}, clinical dialogue~\cite{shi-etal-2024-medical,wu2024medkp}, and question answering~\cite{zhang2023pmc,lin2023medical}, without spatially localization predictions. Subsequent advances~\cite{sellergren2025medgemma, deperrois2025radvlm, bannur2024maira} incorporate stronger multimodal backbones ~\cite{team2025gemma,li2024llava, liu2023visual} and more comprehensive medical supervision. While these models enhance cross-modal alignment and downstream performance, only a subset explicitly incorporates objectives for abnormality localization via detection-style outputs~\cite{deperrois2025radvlm, bannur2024maira}. 

Besides, existing medical VLMs suffer from large sizes and substantial demands for annotation and computation, constraining their applicability in resource-limited environments. Besides, they often treat diagnostic terms as standalone entities or sentence-level report inputs without decomposing them into interpretable visual attributes, which hinders their ability to handle long-tailed, spatially complex, and semantically ambiguous findings, especially in out-of-distribution and compositional scenarios. These challenges reveal the need for efficient and lightweight grounding-oriented models.

\noindent\textbf{Knowledge Injection in CLIP-based Pretraining.} Knowledge injection~\cite{hu2023survey, shrestha2023medical} has emerged as an effective strategy for improving CLIP-style vision-language pretraining, particularly within the medical domain. These approaches commonly rely on contrastive image-text encoder architectures~\cite{radford2021learning, gao2024softclip}, where domain-specific knowledge or enriched prompts are introduced during pretraining. The resulting models are typically evaluated on downstream tasks such as classification~\cite{wang2022multi, zhang2023knowledge, wu2023medklip, chen2022align, luo2024devide}.
For instance, Wang et al.~\cite{wang2022multi} demonstrate that integrating radiology reports and designing multi-granularity supervision during pretraining enhances performance on classification and detection tasks when followed by task-specific fine-tuning. Other methods, such as MedKLIP~\cite{wu2023medklip} and KAD~\cite{zhang2023knowledge}, adopt contrastive similarity for zero-shot classification. 
% However, they are not designed for direct application to detection tasks.
However, these models are primarily optimized for multi-modal pretraining and are not designed for direct localization tasks. Thus, existing knowledge-injected models typically rely on frozen image encoders or lightweight linear probes with further adapt to detection backbone to do localization tasks, they fall short in supporting the fine-grained localization and structured reasoning required by our grounding-based setup. 

In contrast, K2Sight is a framework that decomposes complex clinical concepts into interpretable visual attributes, which enable general-purpose vision-language models to directly perform abnormality grounding. This design allows compact models to achieve performance comparable to current specialized medical VLMs, while requiring fewer computational resources and less supervision.

\begin{figure*}[!t]
\centerline{\includegraphics[width=\textwidth]{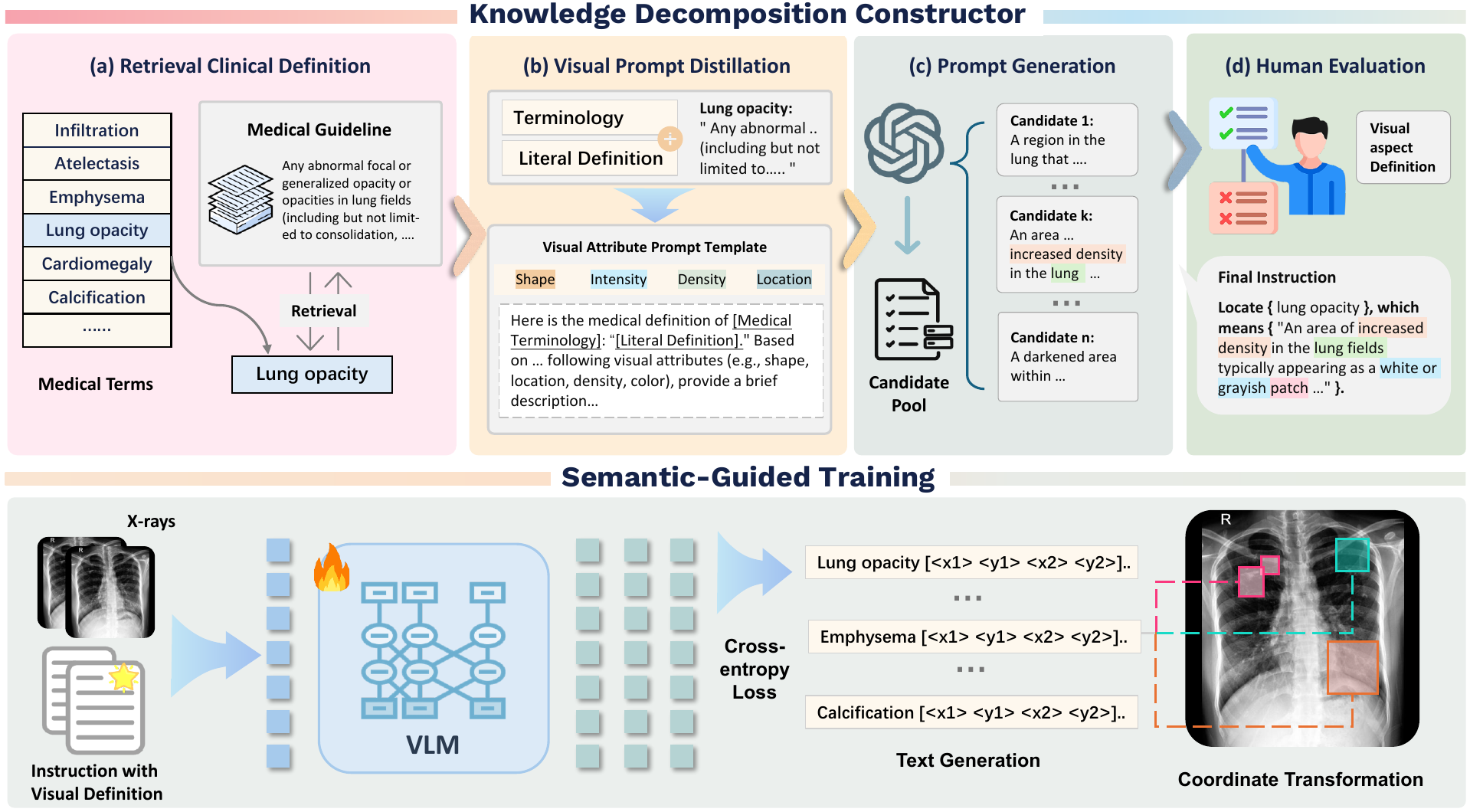}}
\caption{\textbf{Overview of the K2Sight framework.} \textbf{Top: Knowledge Decomposition.} Clinical definitions are retrieved and decomposed into shape, intensity, density and location as core visual attributes. A large language model generates attribute-specific prompts, and human evaluation selects the most faithful and discriminative ones. \textbf{Bottom: Semantic-Guided Training.} Each image is paired with the selected prompts and used to train a vision-language model for abnormality grounding.}
\label{fig:method}
\end{figure*}

\section{Method}
\label{sec:method}
We propose \textbf{K2Sight}, a two-stage framework that addresses \textit{abnormality grounding} through structured knowledge decomposition and vision-language alignment.
K2Sight operates through two stages, as illustrated in Figure~\ref{fig:method}:  
\textit{(1) \textbf{Knowledge Decomposition Constructor}} (Sec.~\ref{sec:decomp}), which performs structured semantic factorization of clinical definitions into attribute-centric visual descriptors;  
\textit{(2) \textbf{Semantic-Guided Training}} (Sec.~\ref{sec:train}), which leverages these decomposed representations for enhanced spatial-semantic alignment through attribute-conditioned learning.
In this framework, we aim to establish a principled mapping between clinical expertise and visual understanding through semantic attribute disentanglement, enabling more effective training of VLMs for abnormality grounding tasks.  

\subsection{Knowledge Decomposition Constructor}
\label{sec:decomp}
A core component of our framework is the conversion of clinical definitions into visually interpretable descriptions. This process enhances abnormality grounding by mapping diagnostic language to attribute-based semantic concepts and is referred to as \textbf{Knowledge to Sight}. It can be formulated as a transformation function $\phi$, which converts the clinical definition $d(a)$ of a radiological abnormality $a$ into a visually grounded instruction $k(a)$.
$$\phi: d(a) \mapsto k(a), \quad a \in \mathcal{A}$$
Here, $\mathcal{A}$ denotes the set of abnormalities under consideration, such as~\emph{“lung opacity"}. The goal is to distill textual medical knowledge $d(a)$ into concise, image-relevant instructions $k(a)$ that encapsulate key visual attributes.

\vspace{0.4em}
\noindent\textbf{(a) Clinical-definition retrieval.}
For each abnormality class $a \in \mathcal{A}$, we retrieve its corresponding clinical definition $d(a)$ from authoritative radiological sources~\cite{nguyen2022vindr, hansell2008fleischner}. These definitions encode expert-level domain knowledge, often combining morphological descriptions with pathophysiological reasoning and diagnostic considerations. For instance, the definition for \textit{lung opacity} is:

\textit{“Any abnormal focal or generalized opacity or opacities in lung fields (including but not limited to consolidation, cavity, fibrosis, nodule, mass, calcification, ...)."}

While rich in semantics, such definitions often contain non-visual or latent clinical concepts that are challenging for current vision-language models to process. The gap between expert-level textual abstractions and image-level perceptual evidence introduces semantic dispersion, limiting their effectiveness for direct grounding.
To address this, we design a targeted distillation mechanism that selectively retains visually actionable elements while filtering out abstract or diagnostically inferential content. This step transforms raw clinical language into a form that is more tractable for visual interpretation and model integration.

\noindent\textbf{(b) Visual Prompt Distillation.} 
To translate textual definitions into visually relevant prompts, we rely on fundamental radiological visual primitives: shape, intensity, density, and anatomical location~\cite{radiopaedia_radiomics,aerts2014radiomics}. These elements serve as the common representational space between expert interpretation and model perception, facilitating structured semantic alignment.
Each definition $d(a)$ is decomposed into two latent components: the target concept (e.g., ``lung opacity'') and its literal clinical description. These are embedded into a carefully constructed prompt template $\pi(a, d(a))$ that instructs a language model to generate a focused visual description. For example:
\begin{tcolorbox}[colback=gray!3!white, colframe=blue!60!black, boxrule=0.8pt, arc=2pt, left=8pt, right=2pt, top=2pt, bottom=2pt, enhanced jigsaw]
\scriptsize
\texttt{\textbf{Prompt:} Here is the medical definition of lung opacity: ``Any abnormal focal or generalized opacity or opacities in lung fields (including but not limited to consolidation, cavity, fibrosis, nodule, mass, calcification, interstitial thickening)...'' Based on this definition, and focusing on shape, intensity, density, and location, provide a concise visual description that could guide image recognition.}
\end{tcolorbox}
The prompt $\pi(a, d(a))$ is provided to a large language model $\mathcal{L}$ to generate a set of candidate visual descriptions:
$$
\{\tilde{k}_i(a)\}_{i=1}^N = \mathcal{L}(\pi(a, d(a)))
$$
This generation step translates expert knowledge into visually grounded prompts in a controlled and semantically structured manner, enabling decomposes complex clinical concepts into visually interpretable semantic primitives.

\noindent\textbf{(c) Prompt Selection with Human Alignment.}  
To ensure clinical accuracy and visual distinctiveness, human annotators review the candidate descriptions and select the one that most faithfully captures the visual manifestation:  
$
k(a) = \text{Select}\left( \{ \tilde{k}_i(a) \}_{i=1}^N \right).
$ 
For example, the final prompt for \textit{lung opacity} is:  
\textit{“An area of increased density in the lung fields, typically appearing as a white or grayish patch.”}  
Given the sensitivity and domain specificity of medical data, we include human verification to ensure semantic accuracy. While the process can be fully automated, manual review helps prevent clinically ambiguous prompts. It involves only high-level validation against definitions, requires no expert annotation, and remains efficient. See supplementary material for details.

\subsection{Semantic-Guided Training}
\label{sec:train}
Based on the decomposed knowledge prompts $k(a)$, we adopt an end-to-end fine-tuning strategy that integrates attribute-level descriptions into a vision-language model to enhance grounding of radiological abnormalities.

\noindent\textbf{Training Objective.} 
The grounding task for VLM is framed as an autoregressive sequence-generation problem over discretized bounding box coordinates. Localization is treated as a conditional token generation process, wherein the model sequentially predicts spatial descriptors conditioned on both visual input and semantic prompts.

Given an input radiograph $I$ and its associated knowledge-enhanced prompt $k(a)$, the model outputs a sequence of tokens that encode the bounding box coordinates. For each abnormality instance, the bounding box is represented as $L_m = [x_0, y_0, x_1, y_1]$, where each coordinate is linearly quantized into a discrete vocabulary $\{0, 1, \ldots, 1000\}$.
Model training follows a standard vision-language learning paradigm, using cross-entropy loss over the autoregressively generated coordinate tokens:
$$
\mathcal{L} = -\sum_{m=1}^{M}\sum_{j=1}^{4} \log p(y_{m,j} \mid I, k(a), y_{m,<j})
$$
Here, $M$ denotes the number of abnormality instances, $y_{m,j}$ is the $j$-th token of the $m$-th bounding box, and $y_{m,<j}$ refers to the previously generated tokens in the sequence. 

During training, the prompt $k(a)$ serves as an inductive bias that guides the model’s attention toward semantically relevant spatial regions and structures. 
This mechanism enables the model to draw from two complementary sources of information. On one hand, it leverages its pretrained understanding of general visual concept patterns. On the other hand, in the presence of semantically diverse and sparsely distributed disease categories, the use of structured prompts introduces consistency within the representation space. By emphasizing shared visual characteristics, the model is encouraged to identify transferable patterns that recur across different conditions. This abstraction process enhances the model’s ability to generalize from limited examples and improves spatial localization in a clinically coherent manner.

\section{Experiments}

\noindent\textbf{Benchmark Baselines.} In the experiment, we evaluate two types of vision language models:
1) Generalist VLMs. These are open-domain models trained on a wide range of vision-language tasks. We assess two recent SOTA series: Qwen2-VL-Instruct~\cite{wang2024qwen2} and InternVL3~\cite{zhu2025internvl3}, both specifically optimized for grounding tasks. Using their publicly available checkpoints, we perform zero-shot evaluation on our datasets to investigate whether current SOTA VLMs can directly handle medical abnormality grounding.
2) Medical-specialist VLMs. These models are fine-tuned on large-scale, domain-specific medical datasets and are capable of performing multiple tasks, including abnormality grounding. We benchmark MAIRA-2~\cite{bannur2024maira} and RadVLM~\cite{deperrois2025radvlm}. MAIRA-2, a 13B model, is trained on 501,825 samples sourced from MIMIC-CXR~\cite{johnson2019mimic}, PadChest~\cite{bustos2020padchest}, and USMix~\cite{demner2016preparing}. RadVLM, a 7B model, is trained on 1,022,742 image-instruction pairs curated from MIMIC-CXR, CheXpert~\cite{irvin2019chexpert}, CheXpert-Plus~\cite{chambon2024chexpert}, MS-CXR~\cite{boecking2022making}, PadChest-GR~\cite{castro2024padchest}, and VinDr-CXR~\cite{nguyen2022vindr}.

\noindent\textbf{Datasets.}
We conduct experiments on two benchmark datasets: VinDr-CXR~\cite{nguyen2022vindr} and PadChest-GR~\cite{castro2024padchest}. Both datasets feature human-annotated bounding box labels provided by expert radiologists.
VinDr-CXR serves as the primary training source and includes annotations for various chest abnormalities. Following prior work~\cite{muller2024chex,deperrois2025radvlm,bannur2024maira}, we apply weighted box fusion~\cite{muller2024chex} to merge overlapping annotations, resulting in 18,195 image–abnormality pairs, split into 16,087 for training and 2,108 for testing.
PadChest-GR is used exclusively for evaluation, and comprises two subsets: one for standard zero-shot evaluation and another for out-of-distribution (OOD) testing. The zero-shot subset contains 641 image–abnormality pairs, while the OOD subset includes 644 pairs without abnormality types overlapping with those in VinDr-CXR. 

\noindent\textbf{Implementation Details.} In the first-stage Knowledge Decomposition Constructor, we utilize GPT-4o~\cite{achiam2023gpt} to generate attribute-specific descriptions. For each abnormality category, we construct a candidate pool of size 5 through multiple sampling rounds. Final instruction sets are selected based on human evaluation to ensure quality and relevance. During generation, the temperature is set to 0.7 and the Top-p value is also fixed at 0.7 to balance creativity and coherence.
In the second-stage Semantic-guided Training, we conduct experiments on two lightweight VLM backbones to validate the effectiveness and generalizability of our proposed K2Sight framework. Specifically, we adopt Florence-2 Base (0.23B)\cite{xiao2024florence} as the backbone for K2Sight Lite, and Qwen2-VL-2B-Instruct\cite{wang2024qwen2} as the backbone for K2Sight Base. Both models are fully fine-tuned in an end-to-end manner. We employ the Adam optimizer with standard hyperparameters, and training is performed on 4×A100 GPUs. Additional details, including data preprocessing and training schedules, are in the supplementary materials.

\noindent \textbf{Evaluation metrics.}
We evaluate all models using standard abnormality detection metrics~\cite{padilla2020survey}, including mean average precision (mAP) at various Intersection over Union (IoU) thresholds: $mAP_{30}$, $mAP_{50}$, $mAP_{75}$, and $mAP_{50}^{95}$.
In addition, we report the Robust Detection Outcome (RoDeO)~\cite{meissen2023robust}, a metric designed for evaluating pathology detection quality. RoDeO comprises three components: classification accuracy ($R_{cls}$), shape matching ($R_{shape}$), and localization precision ($R_{loc}$), which are aggregated into an overall score $R_{total}$.

\subsection{Comparison Results} 
\begin{table}[t]
\renewcommand{\arraystretch}{1.1}
  \centering
  \caption{Baseline performance of generalist VLMs on abnormality grounding tasks using VinDr-CXR and PadChest-GR. All models are evaluated using publicly available checkpoints. Metrics are mAP at IoU thresholds (\%). \textbf{Takeaway:} Generalist VLMs still struggle with abnormality grounding in the medical domain.}
  \begin{adjustbox}{max width=\linewidth}
  \begin{tabular}{l | ccc | ccc}
    \toprule
    \textbf{Model} & \multicolumn{3}{c|}{VinDr-CXR} & \multicolumn{3}{c}{PadChest-GR} \\
    & mAP$_{30}$ & mAP$_{50}$ & mAP$_{75}$ & mAP$_{30}$ & mAP$_{50}$ & mAP$_{75}$ \\
    \midrule
    Qwen2-VL-7B & \textbf{1.52} & \textbf{0.48} & \textbf{0.02} & \textbf{1.20} & \textbf{0.27} & \textbf{0.09} \\
    Qwen2-VL-2B & 0.10 & 0.04 & 0.00 & 0.26 & 0.07 & 0.01 \\
    \midrule
    InternVL3-8B & 0.24 & 0.04 & 0.00 & 0.40 & 0.08 & 0.00 \\
    InternVL3-2B & 0.14 & 0.01 & 0.00 & 0.27 & 0.12 & 0.00 \\
    
    \bottomrule
  \end{tabular}
  \end{adjustbox}
  \label{tab:general_vlm_performancel}
\end{table}

\noindent \textit{\textbf{1) Can generalist VLMs sufficiently perform abnormality grounding?}}
As shown in Table~\ref{tab:general_vlm_performancel}, grounding accuracy remains consistently low across both datasets. 
Qwen2-VL-7B Instruct achieves the highest scores among evaluated models, yet its overall performance remains limited.
The best result reaches only 1.52\% mAP$_{30}$ and 0.48\%  mAP$_{50}$. 
Some models exhibit mAP${75}$ close to 0.0, reflecting a complete failure in fine-grained grounding. 
We further evaluate generalist model performance using optimized visual-oriented descriptions generated by our framework. Although certain metrics show slight improvement, overall accuracy remains comparably low (see supplementary materials).

These findings suggest that increasing model size alone does not overcome the challenges of abnormality grounding, and generalist VLMs face clear limitations in clinical localization tasks.
\begin{table*}[t]
  \centering
  \small
\caption{Comparison of abnormality grounding performance (\%) on the VinDr-CXR and PadChest-GR datasets. All metrics are reported as percentages. $^{\dagger}$ denotes zero-shot evaluation on PadChest-GR. $R_\text{loc}$, $R_\text{shape}$, $R_\text{cls}$, and $R_\text{total}$ represent the location, shape, classification, and overall values of the RoDeO metrics, respectively. The best results for each metric are shown in \textbf{bold}, and the second-best in \underline{underlined}. \textbf{Takeaway:} K2Sight delivers competitive or superior performance across most metrics compared to state-of-the-art, medical-specialist VLMs, while using significantly fewer parameters and less supervision.}

  \label{tab:main_comp}
  \begin{adjustbox}{width=\linewidth}
    \begin{tabular}{l l l >{\centering\arraybackslash}p{0.7cm} *{7}{>{\centering\arraybackslash}p{1.2cm}}}
      \toprule
      \textbf{Dataset} & \textbf{Model} & \textbf{Backbone} & \textbf{Param.} & $\text{mAP}_{50}^{95} \uparrow$ & $\text{mAP}_{50} \uparrow$ & $\text{mAP}_{75} \uparrow$ & $R_\text{loc} \uparrow$ & $R_\text{shape} \uparrow$ & $R_\text{cls} \uparrow$ & $R_\text{total} \uparrow$ \\

      \midrule
      \multirow{4}{*}{VinDr-CXR} 
        & RadVLM~\cite{deperrois2025radvlm}      & LLaVA-OneVision & 7B     & 8.62 & 21.86 & 5.78 & 55.96 & 38.50 & \underline{73.60} & 52.24 \\
        & MAIRA-2~\cite{bannur2024maira}         & LLaVA            & 13B    & 1.22 & 4.92  & 0.32 & 27.92 & 17.23 & 56.56 & 26.90 \\

        & \cellcolor{gray!15}K2Sight-Light    & \cellcolor{gray!15}Florence-2 & \cellcolor{gray!15}0.23B  
        & \cellcolor{gray!15}\textbf{9.93} & \cellcolor{gray!15}\underline{23.78} & \cellcolor{gray!15}\textbf{7.60} 
        & \cellcolor{gray!15}\textbf{61.83} & \cellcolor{gray!15}\textbf{46.88} & \cellcolor{gray!15}73.02 & \cellcolor{gray!15}\textbf{58.59} \\

        & \cellcolor{gray!15}K2Sight-Base      & \cellcolor{gray!15}Qwen2-VL   & \cellcolor{gray!15}2B     
        & \cellcolor{gray!15}\underline{9.13} & \cellcolor{gray!15}\textbf{24.22} & \cellcolor{gray!15}\underline{6.57} 
        & \cellcolor{gray!15}\underline{59.66} & \cellcolor{gray!15}\underline{40.87} & \cellcolor{gray!15}\textbf{77.35} & \cellcolor{gray!15}\underline{55.40} \\

      \midrule
        \multirow{4}{*}{PadChest-GR} 
        & RadVLM~\cite{deperrois2025radvlm}      & LLaVA-OneVision & 7B     
        & 2.40 & 10.55 & 0.36 
        & \underline{48.23} & 24.39 & 72.25 & 39.69 \\
        
        & MAIRA-2~\cite{bannur2024maira}         & LLaVA            & 13B    
        & \textbf{8.34} & \textbf{19.19} & \textbf{5.70} 
        & 22.78 & 20.37 & 40.19 & 25.45 \\
        
        & \cellcolor{gray!15}K2Sight-Light$^{\dagger}$ 
        & \cellcolor{gray!15}Florence-2 & \cellcolor{gray!15}0.23B  
        & \cellcolor{gray!15}2.94 & \cellcolor{gray!15}11.71 & \cellcolor{gray!15}0.35 
        & \cellcolor{gray!15}42.81 & \cellcolor{gray!15}\underline{26.57} & \cellcolor{gray!15}\textbf{74.53} & \cellcolor{gray!15}\underline{40.31} \\
        
        & \cellcolor{gray!15}K2Sight-Base$^{\dagger}$   
        & \cellcolor{gray!15}Qwen2-VL   & \cellcolor{gray!15}2B     
        & \cellcolor{gray!15}\underline{3.21} & \cellcolor{gray!15}\underline{13.14} & \cellcolor{gray!15}\underline{0.58} 
        & \cellcolor{gray!15}\textbf{48.28} & \cellcolor{gray!15}\textbf{26.71} & \cellcolor{gray!15}\underline{72.54} & \cellcolor{gray!15}\textbf{41.70} \\
        
      \bottomrule
    \end{tabular}
  \end{adjustbox}
\end{table*}

% =======================2=============================

\noindent\textit{\textbf{2) Can K2Sight lift compact, general-purpose VLMs to specialist level?}}  
\label{sec:comparison}
Table~\ref{tab:main_comp} benchmarks K2Sight-Lite (0.23B) and K2Sight-Base (2B) against the radiology-specific RadVLM and MAIRA-2 on VinDr-CXR and PadChest-GR. RadVLM and MAIRA-2 are pre-trained on approximately 1 million and 0.5 million radiology samples, respectively. VinDr-CXR images (constituting roughly 1.6\% of RadVLM’s training data) are used for training, while evaluation on PadChest-GR is conducted in a zero-shot setting.

On VinDr-CXR dataset, K2Sight-Lite achieves the highest $\text{mAP}_{50}^{95}$ at 9.93\%, $\text{mAP}_{75}$ at 7.60\%, while K2Sight-Base leads in $\text{mAP}_{50}$ with 24.22\% and in $R_\text{cls}$ with 77.35\%, highlighting the advantage of decomposed localization and classification supervision.  
Despite having significantly fewer parameters, K2Sight-Lite remains competitive, which may be attributed to the Florence-2 backbone that enhances spatial representations.  

In the zero-shot evaluation on PadChest-GR, K2Sight-Base achieves the highest scores in $R_\text{loc}$, $R_\text{shape}$, and the overall metric $R_\text{total}$, reaching 41.70\%. Additionally, both K2Sight variants outperform RadVLM in $R_\text{cls}$, with scores of 74.53\% and 72.54\%, respectively.

Although MAIRA-2 reports higher $\text{mAP}$ values on PadChest-GR, its performance benefits from fine-tuning on that dataset, making the comparison less indicative of generalization.
These results demonstrate that structured supervision and visually grounded prompts enable general-purpose VLMs to achieve specialist-level performance under limited data. This approach provides a scalable solution for developing efficient, generalizable, and clinically deployable medical VLMs.

\subsection{Ablation Study}
\label{sec:ablation1}

\definecolor{gaincolor}{RGB}{0, 140, 0}
\begin{table*}[t]
  \centering
  \small
  \caption{Ablation study on VinDr-CXR and PadChest-GR datasets, evaluating the impact of our K2Sight framework. Settings marked as “w/ KD” correspond to our full K2Sight pipeline, integrating visual definition prompts and knowledge distillation. “Base” refers to models trained using the same backbone and supervision setup, but without visual attribute prompts or Knowledge Decomposition Constructor (KDC). ∆ rows represent absolute gains from w/o KDC to w/ KDC. \textbf{Takeaway:} Decomposing complex medical knowledge via K2Sight consistently enhances grounding performance across datasets and model scales.}
  \label{tab:ablation_study_updated}
  \begin{adjustbox}{width=\linewidth}
  \begin{tabular}{l l l >{\centering\arraybackslash}p{1.2cm} *{7}{>{\centering\arraybackslash}p{1.2cm}}}
    \toprule
    \textbf{Dataset} & \textbf{Model} & \textbf{Setting} & \textbf{Params} & $\text{mAP}_{50}^{95} \uparrow$ & $\text{mAP}_{50} \uparrow$ & $\text{mAP}_{75} \uparrow$ & $R_\text{loc} \uparrow$ & $R_\text{shape} \uparrow$ & $R_\text{cls} \uparrow$ & $R_\text{total} \uparrow$ \\
    \midrule
    \multirow{6}{*}{VinDr-CXR}
      & Florence-2 (Base)         & w/o KDC & 0.23B & 5.84 & 13.96 & 5.00 & 44.70 & 33.29 & 71.14 & 45.14 \\
      & K2Sight-Light             & w/ KDC  & 0.23B & \textbf{9.93} & \textbf{23.78} & \textbf{7.60} & \textbf{61.83} & \textbf{46.88} & \textbf{73.02} & \textbf{58.59} \\
      & \cellcolor{gray!15}∆ (Gain) & \cellcolor{gray!15}— & \cellcolor{gray!15}— 
        & \cellcolor{gray!15}\textcolor{gaincolor}{+4.09} 
        & \cellcolor{gray!15}\textcolor{gaincolor}{+9.82} 
        & \cellcolor{gray!15}\textcolor{gaincolor}{+2.60} 
        & \cellcolor{gray!15}\textcolor{gaincolor}{+17.13} 
        & \cellcolor{gray!15}\textcolor{gaincolor}{+13.59} 
        & \cellcolor{gray!15}\textcolor{gaincolor}{+1.88} 
        & \cellcolor{gray!15}\textcolor{gaincolor}{+13.45} \\
      & Qwen2-VL (Base)           & w/o KDC & 2B & 8.13 & 20.03 & 5.15 & 54.42 & 36.80 & 70.58 & 50.23 \\
      & K2Sight-Base              & w/ KDC  & 2B & \textbf{9.13} & \textbf{24.22} & \textbf{6.57} & \textbf{59.66} & \textbf{40.87} & \textbf{77.35} & \textbf{55.40} \\
      & \cellcolor{gray!15}∆ (Gain) & \cellcolor{gray!15}— & \cellcolor{gray!15}— 
        & \cellcolor{gray!15}\textcolor{gaincolor}{+1.00} 
        & \cellcolor{gray!15}\textcolor{gaincolor}{+4.19} 
        & \cellcolor{gray!15}\textcolor{gaincolor}{+1.42} 
        & \cellcolor{gray!15}\textcolor{gaincolor}{+5.24} 
        & \cellcolor{gray!15}\textcolor{gaincolor}{+4.07} 
        & \cellcolor{gray!15}\textcolor{gaincolor}{+6.77} 
        & \cellcolor{gray!15}\textcolor{gaincolor}{+5.17} \\
    \midrule
    \multirow{6}{*}{PadChest-GR$^{\dagger}$}
      & Florence-2 (Base)         & w/o KDC & 0.23B & 2.25 & 8.90 & \textbf{0.48} & 39.85 & 26.20 & 73.16 & 39.00 \\
      & K2Sight-Light$^{\dagger}$ & w/ KDC  & 0.23B & \textbf{2.94} & \textbf{11.71} & 0.35 & \textbf{42.81} & \textbf{26.57} & \textbf{74.53} & \textbf{40.31} \\
      & \cellcolor{gray!15}∆ (Gain) & \cellcolor{gray!15}— & \cellcolor{gray!15}— 
        & \cellcolor{gray!15}\textcolor{gaincolor}{+0.69} 
        & \cellcolor{gray!15}\textcolor{gaincolor}{+2.81} 
        & \cellcolor{gray!15}{-0.13} 
        & \cellcolor{gray!15}\textcolor{gaincolor}{+2.96} 
        & \cellcolor{gray!15}\textcolor{gaincolor}{+0.37} 
        & \cellcolor{gray!15}\textcolor{gaincolor}{+1.37} 
        & \cellcolor{gray!15}\textcolor{gaincolor}{+1.31} \\
      & Qwen2-VL (Base)           & w/o KDC & 2B & 3.14 & 12.37 & \textbf{0.80} & 43.71 & 24.77 & 67.85 & 38.47 \\
      & K2Sight-Base$^{\dagger}$  & w/ KDC  & 2B & \textbf{3.21} & \textbf{13.14} & 0.58 & \textbf{48.28} & \textbf{26.71} & \textbf{72.54} & \textbf{41.70} \\
      & \cellcolor{gray!15}∆ (Gain) & \cellcolor{gray!15}— & \cellcolor{gray!15}— 
        & \cellcolor{gray!15}\textcolor{gaincolor}{+0.07} 
        & \cellcolor{gray!15}\textcolor{gaincolor}{+0.77} 
        & \cellcolor{gray!15}{-0.22} 
        & \cellcolor{gray!15}\textcolor{gaincolor}{+4.57} 
        & \cellcolor{gray!15}\textcolor{gaincolor}{+1.94} 
        & \cellcolor{gray!15}\textcolor{gaincolor}{+4.69} 
        & \cellcolor{gray!15}\textcolor{gaincolor}{+3.23} \\
    \bottomrule
  \end{tabular}
  \end{adjustbox}
\end{table*}

\definecolor{dropMax}{RGB}{69,117,180}   
\definecolor{dropHigh}{RGB}{116,173,209}
\definecolor{dropMid}{RGB}{171,217,233}
\definecolor{dropLow}{RGB}{224,243,248}

\newcommand{\colmax}{\cellcolor{dropMax}}
\newcommand{\colhigh}{\cellcolor{dropHigh}}
\newcommand{\colmid}{\cellcolor{dropMid}}
\newcommand{\collow}{\cellcolor{dropLow}}

\begin{table}[!t]
\centering
\caption{Attribute-conditioned ablation study on VinDr-CXR using K2Sight light and base model. At inference time, each column masks a single visual attribute to isolate its contribution. Absolute metric values are reported, with changes relative to full prompting shown in parentheses. \textbf{Takeaway:} All attributes affect grounding performance; intensity is comparatively less contribution, and the base model exhibits lower sensitivity than light version.}
\small
\begin{adjustbox}{width=\columnwidth}
\begin{tabular}{l|c|c|c|c|c}
\toprule
\textbf{Metric} & \textbf{Base} & \textbf{w/o Shape} & \textbf{w/o Intensity} & \textbf{w/o Density} & \textbf{w/o Location} \\
\midrule
\multicolumn{6}{c}{\textbf{K2Sight-Light}} \\
\midrule
mAP$_{50}^{95}$ & 9.93 & \collow 7.96 (-1.97) & \collow 8.28 (-1.65) & \colmid 7.82 (-2.11) & \colmid 7.76 (-2.17) \\
mAP$_{50}$      & 23.78 & \colhigh 18.26 (-5.52) & \colmid 19.03 (-4.75) & \colhigh 17.95 (-5.83) & \colhigh 18.16 (-5.62) \\
mAP$_{75}$      & 7.60 & \collow 6.15 (-1.45) & \collow 6.21 (-1.39) & \collow 5.82 (-1.78) & \collow 6.05 (-1.55) \\
R$_\text{loc}$  & 61.83 & \colmax 53.55 (-8.28) & \colhigh 53.98 (-7.85) & \colmax 52.34 (-9.49) & \colmax 53.01 (-8.82) \\
R$_\text{shape}$ & 46.88 & \colhigh 41.20 (-5.68) & \colhigh 41.25 (-5.63) & \colhigh 40.50 (-6.38) & \colhigh 40.20 (-6.68) \\
R$_\text{cls}$  & 73.02 & \colhigh 67.34 (-5.68) & \colmid 68.22 (-4.80) & \colmax 63.41 (-9.61) & \colmid 69.11 (-3.91) \\
R$_\text{total}$ & 58.59 & \colhigh 51.91 (-6.68) & \colhigh 52.24 (-6.35) & \colmax 50.36 (-8.23) & \colhigh 51.54 (-7.05) \\
\midrule
\multicolumn{6}{c}{\textbf{K2Sight-Base}} \\
\midrule
mAP$_{50}^{95}$ & 9.13 & \collow 8.58 (-0.55) & \collow 7.73 (-1.40) & \collow 7.27 (-1.86) & \collow 7.80 (-1.33) \\
mAP$_{50}$      & 24.22 & \collow 24.36 (+0.14) & \collow 22.88 (-1.34) & \colmid 21.47 (-2.75) & \colmid 22.01 (-2.21) \\
mAP$_{75}$      & 6.57 & \colmid 3.43 (-3.14) & \colmid 3.50 (-3.07) & \colmid 3.46 (-3.11) & \colmid 3.85 (-2.72) \\
R$_\text{loc}$  & 59.66 & \colmid 56.37 (-3.29) & \colmid 55.70 (-3.96) & \collow 58.49 (-1.17) & \colmid 57.35 (-2.31) \\
R$_\text{shape}$ & 40.87 & \colmid 38.20 (-2.67) & \colmid 37.58 (-3.29) & \collow 39.37 (-1.50) & \colmid 38.72 (-2.15) \\
R$_\text{cls}$  & 77.35 & \colmid 73.21 (-4.14) & \colhigh 72.24 (-5.11) & \collow 75.88 (-1.47) & \colmid 74.32 (-3.03) \\
R$_\text{total}$ & 55.40 & \colmid 52.11 (-3.29) & \colmid 51.36 (-4.04) & \collow 53.88 (-1.52) & \colmid 52.89 (-2.51) \\
\bottomrule
\end{tabular}
\end{adjustbox}
\label{tab:attr_ablation_colored}
\end{table}

\noindent\textit{\textbf{1) How effective is the K2Sight framework?}}  
We evaluate the central contribution of the K2Sight framework, which introduces Knowledge Decomposition Constructor (KDC) with visual attribute-based supervision. Table~\ref{tab:ablation_study_updated} compares models trained with our supervision strategy against counterparts using identical backbones and fine-tuning pipelines but without decomposed attributes or visual prompts.

On VinDr-CXR, applying KDC with the Florence-2 backbone increases $\text{mAP}_{50}$ by 9.82~\% points, $R_\text{loc}$ by 17.13~\% points, and $R_\text{shape}$ by 13.59~\% points, leading to a 13.45~\% point gain in the overall $R_\text{total}$.
With Qwen2-VL, KDC improves $\text{mAP}_{50}$ by 4.19~\% points and lifts $R\text{cls}$ by 6.77~\% points, despite the model’s larger scale and pretraining capacity.
These consistent gains suggest that attribute decomposition complements both compact and large vision-language models by explicitly guiding localization and classification learning.

In the zero-shot setting on PadChest-GR, K2Sight continues to yield considerable benefits. Florence-2 gains 2.81\% points in $\text{mAP}_{50}$ and 2.96~\% in $R_\text{loc}$. Qwen2-VL also improves in $R_\text{loc}$ and $R_\text{cls}$, resulting in an overall increase of $R_\text{total}$ by 3.23~\% points.
Minor drops in $\text{mAP}_{75}$, such as a 0.13~\% point decrease for Florence-2, suggest that while high-IoU precision may slightly decline, the overall grounding coverage and robustness are significantly improved.

These findings show that visual attribute-based supervision improves abnormality grounding and generalizes across datasets and model architectures. Despite limited training data, KDC enhances both spatial and semantic representation learning, demonstrating its effectiveness as a data-efficient strategy for clinical localization. 

\noindent\textit{\textbf{2) How does each attribute affect performance?}}  
We perform attribute-conditioned ablation during inference to assess the individual impact of each visual prompt component. As shown in Table~\ref{tab:attr_ablation_colored}, masking any single attribute leads to performance degradation, confirming that all components contribute to effective grounding.

K2Sight-Light exhibits higher sensitivity compared to the K2Sight-Base. 
For K2Sight-Light, masking the density attributes leads to the worth degradation in performance, with $R_\text{loc}$ dropping by 9.49~\% points.
Location is similarly impactful, with an 8.82~\% point drop in $R_\text{loc}$ and 7.05~\% points in $R_\text{total}$. Shape and intensity have a more moderate influence, although masking shape still reduces $\text{mAP}_{50}$ by over 5~\% points. Intensity consistently shows the smallest effect across all metrics, indicating that this attribute is either less informative or more easily compensated for by the model. Regarding K2Sight-Base, when density masking is applied, $R_\text{loc}$ drops by only 1.17~\% points and $R_\text{cls}$ by 1.47~\% points and $R_\text{cls}$ by 1.47~\% points. Masking shape or intensity leads to negligible changes in $\text{mAP}_{50}$, suggesting that the larger model has stronger internal representations and is more robust to prompt perturbations. Nevertheless, for both light and base variants, $\text{mAP}_{75}$ drops by more than 3~\% points across all ablations, indicating a shared reliance on full attribute context for achieving high-IoU precision.

These results reveal that spatial and structural attributes like location and density play a major role in guiding the grounding process. Their removal disproportionately impacts localization and classification performance. The disparity in sensitivity between light and base variants suggests that smaller models rely more heavily on explicit attribute to compensate for limited capacity, whereas larger models can internalize these cues more effectively. For more details on the masking strategy, please refer to the supplementary materials.

\subsection{Further Analysis}

\begin{table}[!t]
\centering
\caption{Evaluation on unseen diseases in the PadChest-GR dataset. F2 denotes Florence-2 (Base), Qwen denotes Qwen2-VL (Base), and K2S-L / K2S-B represent our proposed K2Sight-Light and K2Sight-Base models. $\Delta$ columns show absolute performance gaps between our models and respective baselines. \textbf{Takeaway:} K2Sight shows stronger generalization to unseen diseases and consistently outperforms generalist baselines.}

\small
\begin{adjustbox}{width=\columnwidth}
\begin{tabular}{l|cc|cc}
\toprule
\textbf{Metric} & \textbf{F2 / K2S-L} &\textbf{ $\Delta$K2S-L} & \textbf{Qwen / K2S-B} & \textbf{$\Delta$K2S-B} \\
\midrule
mAP$_{50}^{95}$   & 0.53 / 0.91 & \textcolor{green!60!black}{+0.38} & 2.82 / 3.03 & \textcolor{green!60!black}{+0.21} \\
mAP$_{50}$        & 1.79 / 3.03 & \textcolor{green!60!black}{+1.24} & 8.91 / 8.54 & \textcolor{red!40!black}{-0.37} \\
mAP$_{75}$        & 0.10 / 0.47 & \textcolor{green!60!black}{+0.37} & 1.01 / 1.16 & \textcolor{green!60!black}{+0.15} \\
R$_\text{loc}$    & 38.14 / 49.99 & \textcolor{green!60!black}{+11.85} & 52.36 / 57.37 & \textcolor{green!60!black}{+5.01} \\
R$_\text{shape}$  & 19.04 / 18.66 & \textcolor{red!40!black}{-0.38} & 25.10 / 27.99 & \textcolor{green!60!black}{+2.89} \\
R$_\text{cls}$    & 69.01 / 77.23 & \textcolor{green!60!black}{+8.22} & 79.39 / 85.04 & \textcolor{green!60!black}{+5.65} \\
R$_\text{total}$  & 32.18 / 34.67 & \textcolor{green!60!black}{+2.49} & 41.94 / 46.22 & \textcolor{green!60!black}{+4.28} \\
\bottomrule
\end{tabular}
\end{adjustbox}
\label{tab:unkown_test}
\end{table}

\begin{figure}[t]
  \centering
  \includegraphics[width=\linewidth,height=9.1cm]{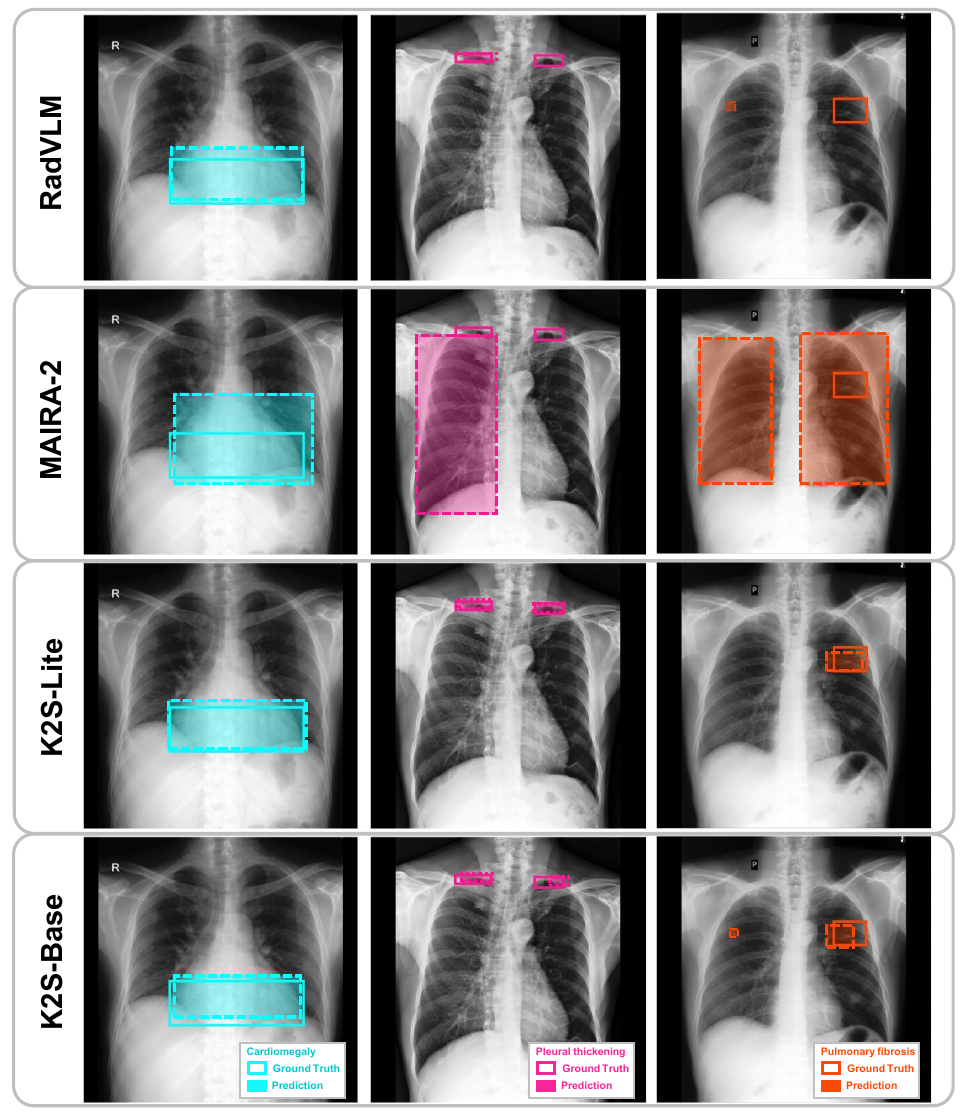}
  \caption{Visualization examples from four models: RadVLM, MAIRA-2, K2Sight-Light, and K2Sight-Base. Ground truth annotations are indicated with solid bounding boxes, while model predictions are visualized using transparent colored overlays. \textbf{Takeaway:} Our K2Sight models consistently demonstrate more accurate and fine-grained localization.}
  \label{fig:examples}
\end{figure}

\noindent\textit{\textbf{1) Can K2Sight generalize to unseen diseases?}}  
To assess generalization performance, we evaluate the K2Sight models on the PadChest-GR-Unknown subset, which includes disease categories that are entirely absent during training. As shown in Table~\ref{tab:unkown_test}, we compare K2Sight-Light and K2Sight-Base against Florence-2 and Qwen2-VL base models, both trained on VinDr-CXR without applying knowledge decomposition, following the same experimental setup described in Section~\ref{sec:ablation1}.

K2Sight consistently outperforms these baselines across multiple grounding metrics. Specifically, K2Sight-Light improves $R_\text{loc}$ by 11.85~\% points and $R_\text{cls}$ by 8.22~\% points compared to Florence-2.
K2Sight-Base also shows considerable improvements, with $R_\text{loc}$ increasing by 5.01~\% points and $R_\text{cls}$ improving by 5.65~\% points over Qwen2-VL. These enhancements are reflected in the overall RoDeo metric $R_\text{total}$, which increases by 4.28~\% points for the base variant.
Although Qwen2-VL achieves a slightly higher $\text{mAP}_{50}$ than K2Sight-Base, with a difference of 0.37~\% points, its overall grounding performance remains inferior.

These findings suggest that training with decomposed, attribute-specific prompts enhances out-of-distribution generalization and grounding fidelity, particularly in clinically diverse settings.

\noindent\textit{\textbf{2) Visualization Example Analysis.}} Figure~\ref{fig:examples} shows the qualitative grounding capabilities of each model. In easy scenarios, such as \say{cardiomegaly}, all models demonstrate reliable localization (first column).
The second column presents a case of \say{pleural thickening}. Both K2Sight-Light and K2Sight-Base accurately localize the two annotated regions. In contrast, RadVLM correctly detects only one region, while MAIRA-2 produces inconsistent predictions that deviate significantly from the ground truth.
The third case focuses on \say{ pulmonary fibrosis}. Although both K2Sight and MAIRA-2 generate bounding boxes that intersect with the ground truth, MAIRA-2's predictions tend to be overly broad. In contrast, K2Sight-Base provides more localized and accurate results. Yet, K2Sight-Base also produces one false positive by predicting an extra bounding box in a normal region.

\section{Conclusion and Discussion}
We propose \textbf{K2Sight}, a novel framework for enhancing abnormality grounding in vision-language models by decomposing complex clinical terminology into semantically visual attributes. This approach enables compact models to exceed the performance of larger general-purpose VLMs, particularly in zero-shot scenarios involving unseen diseases.
Experimental results demonstrate that attribute-based prompting significantly improves both in-domain localization and out-of-distribution generalization, underscoring K2Sight’s effectiveness in data-scarce settings. Its high accuracy, compact design, and interpretability make it well-suited for integration into agentic diagnostic systems, effectively bridging clinical language and visual evidence to support real-world medical AI applications.
% \clearpage %if without this will have an empty line
{
    \small
    \bibliographystyle{ieeenat_fullname}
    \bibliography{main}

\begin{thebibliography}{56}
\providecommand{\natexlab}[1]{#1}
\providecommand{\url}[1]{\texttt{#1}}
\expandafter\ifx\csname urlstyle\endcsname\relax
  \providecommand{\doi}[1]{doi: #1}\else
  \providecommand{\doi}{doi: \begingroup \urlstyle{rm}\Url}\fi

\bibitem[Achiam et~al.(2023)Achiam, Adler, Agarwal, Ahmad, Akkaya, Aleman, Almeida, Altenschmidt, Altman, Anadkat, et~al.]{achiam2023gpt}
Josh Achiam, Steven Adler, Sandhini Agarwal, Lama Ahmad, Ilge Akkaya, Florencia~Leoni Aleman, Diogo Almeida, Janko Altenschmidt, Sam Altman, Shyamal Anadkat, et~al.
\newblock {GPT-4} technical report.
\newblock \emph{arXiv preprint arXiv:2303.08774}, 2023.

\bibitem[Aerts et~al.(2014)Aerts, Velazquez, Leijenaar, Parmar, Grossmann, Carvalho, Bussink, Monshouwer, Haibe-Kains, and Lambin]{aerts2014radiomics}
Hugo~JWL Aerts, Emmanuel~Rios Velazquez, Ralph~TH Leijenaar, Chintan Parmar, Patrick Grossmann, Susana Carvalho, Johan Bussink, Rene Monshouwer, Benjamin Haibe-Kains, and Philippe Lambin.
\newblock Decoding tumour phenotype by noninvasive imaging using a quantitative radiomics approach.
\newblock \emph{Nature communications}, 5:\penalty0 4006, 2014.

\bibitem[Alkhaldi et~al.(2024)Alkhaldi, Alnajim, Alabdullatef, Alyahya, Chen, Zhu, Alsinan, and Elhoseiny]{alkhaldi2024minigpt}
Asma Alkhaldi, Raneem Alnajim, Layan Alabdullatef, Rawan Alyahya, Jun Chen, Deyao Zhu, Ahmed Alsinan, and Mohamed Elhoseiny.
\newblock Minigpt-med: Large language model as a general interface for radiology diagnosis.
\newblock \emph{arXiv preprint arXiv:2407.04106}, 2024.

\bibitem[Arora et~al.(2025)Arora, Wei, Hicks, Bowman, Qui{\~n}onero-Candela, Tsimpourlas, Sharman, Shah, Vallone, Beutel, et~al.]{arora2025healthbench}
Rahul~K Arora, Jason Wei, Rebecca~Soskin Hicks, Preston Bowman, Joaquin Qui{\~n}onero-Candela, Foivos Tsimpourlas, Michael Sharman, Meghan Shah, Andrea Vallone, Alex Beutel, et~al.
\newblock Healthbench: Evaluating large language models towards improved human health.
\newblock \emph{arXiv preprint arXiv:2505.08775}, 2025.

\bibitem[Bai et~al.(2023)Bai, Bai, Yang, Wang, Tan, Wang, Lin, Zhou, and Zhou]{bai2023qwen}
Jinze Bai, Shuai Bai, Shusheng Yang, Shijie Wang, Sinan Tan, Peng Wang, Junyang Lin, Chang Zhou, and Jingren Zhou.
\newblock Qwen-vl: A frontier large vision-language model with versatile abilities.
\newblock \emph{arXiv preprint arXiv:2308.12966}, 2023.

\bibitem[Bannur et~al.(2024)Bannur, Bouzid, Castro, Schwaighofer, Thieme, Bond-Taylor, Ilse, P{\'e}rez-Garc{\'\i}a, Salvatelli, Sharma, et~al.]{bannur2024maira}
Shruthi Bannur, Kenza Bouzid, Daniel~C Castro, Anton Schwaighofer, Anja Thieme, Sam Bond-Taylor, Maximilian Ilse, Fernando P{\'e}rez-Garc{\'\i}a, Valentina Salvatelli, Harshita Sharma, et~al.
\newblock Maira-2: Grounded radiology report generation.
\newblock \emph{arXiv preprint arXiv:2406.04449}, 2024.

\bibitem[Boecking et~al.(2022)Boecking, Usuyama, Bannur, Castro, Schwaighofer, Hyland, Wetscherek, Naumann, Nori, Alvarez-Valle, et~al.]{boecking2022making}
Benedikt Boecking, Naoto Usuyama, Shruthi Bannur, Daniel~C Castro, Anton Schwaighofer, Stephanie Hyland, Maria Wetscherek, Tristan Naumann, Aditya Nori, Javier Alvarez-Valle, et~al.
\newblock Making the most of text semantics to improve biomedical vision--language processing.
\newblock In \emph{European Conference on Computer Vision}, pages 1--21. Springer, 2022.

\bibitem[Bustos et~al.(2020)Bustos, Pertusa, Salinas, and De~La Iglesia-Vaya]{bustos2020padchest}
Aurelia Bustos, Antonio Pertusa, Jose-Maria Salinas, and Maria De~La Iglesia-Vaya.
\newblock Padchest: A large chest x-ray image dataset with multi-label annotated reports.
\newblock \emph{Medical Image Analysis}, 66:\penalty0 101797, 2020.

\bibitem[Castro et~al.(2024)Castro, Bustos, Bannur, Hyland, Bouzid, Wetscherek, S{\'a}nchez-Valverde, Jaques-P{\'e}rez, P{\'e}rez-Rodr{\'\i}guez, Takeda, et~al.]{castro2024padchest}
Daniel~C Castro, Aurelia Bustos, Shruthi Bannur, Stephanie~L Hyland, Kenza Bouzid, Maria~Teodora Wetscherek, Maria~Dolores S{\'a}nchez-Valverde, Lara Jaques-P{\'e}rez, Lourdes P{\'e}rez-Rodr{\'\i}guez, Kenji Takeda, et~al.
\newblock Padchest-gr: A bilingual chest {X-ray} dataset for grounded radiology report generation.
\newblock \emph{arXiv preprint arXiv:2411.05085}, 2024.

\bibitem[Chambon et~al.(2024)Chambon, Delbrouck, Sounack, Huang, Chen, Varma, Truong, Chuong, and Langlotz]{chambon2024chexpert}
Pierre Chambon, Jean-Benoit Delbrouck, Thomas Sounack, Shih-Cheng Huang, Zhihong Chen, Maya Varma, Steven~QH Truong, Chu~The Chuong, and Curtis~P Langlotz.
\newblock {CheXpert Plus}: Augmenting a large chest {X-ray} dataset with text radiology reports, patient demographics and additional image formats.
\newblock \emph{arXiv preprint arXiv:2405.19538}, 2024.

\bibitem[Chen et~al.(2022)Chen, Li, and Wan]{chen2022align}
Zhihong Chen, Guanbin Li, and Xiang Wan.
\newblock Align, reason and learn: Enhancing medical vision-and-language pre-training with knowledge.
\newblock In \emph{Proceedings of the 30th ACM international conference on multimedia}, pages 5152--5161, 2022.

\bibitem[Chen et~al.(2024)Chen, Varma, et~al.]{chen2024chexagent}
Zhihong Chen, Varma, et~al.
\newblock Chexagent: Towards a foundation model for chest {X-ray} interpretation.
\newblock \emph{arXiv preprint arXiv:2401.12208}, 2024.

\bibitem[Demner-Fushman et~al.(2016)Demner-Fushman, Kohli, Rosenman, Shooshan, Rodriguez, Antani, Thoma, and McDonald]{demner2016preparing}
Dina Demner-Fushman, Marc~D Kohli, Marc~B Rosenman, Sonya~E Shooshan, Laritza Rodriguez, Sameer Antani, George~R Thoma, and Clement~J McDonald.
\newblock Preparing a collection of radiology examinations for distribution and retrieval.
\newblock \emph{Journal of the American Medical Informatics Association}, 23\penalty0 (2):\penalty0 304--310, 2016.

\bibitem[Deperrois et~al.(2025)Deperrois, Matsuo, Ruip{\'e}rez-Campillo, Vandenhirtz, Laguna, Ryser, Fujimoto, Nishio, Sutter, Vogt, et~al.]{deperrois2025radvlm}
Nicolas Deperrois, Hidetoshi Matsuo, Samuel Ruip{\'e}rez-Campillo, Moritz Vandenhirtz, Sonia Laguna, Alain Ryser, Koji Fujimoto, Mizuho Nishio, Thomas~M Sutter, Julia~E Vogt, et~al.
\newblock {RadVLM}: A multitask conversational vision-language model for radiology.
\newblock \emph{arXiv preprint arXiv:2502.03333}, 2025.

\bibitem[Gao et~al.(2024)Gao, Liu, Xu, Wu, Zhang, Li, Yang, Liu, and Sun]{gao2024softclip}
Yuting Gao, Jinfeng Liu, Zihan Xu, Tong Wu, Enwei Zhang, Ke Li, Jie Yang, Wei Liu, and Xing Sun.
\newblock Softclip: Softer cross-modal alignment makes clip stronger.
\newblock In \emph{Proceedings of the AAAI Conference on Artificial Intelligence}, pages 1860--1868, 2024.

\bibitem[Hansell et~al.(2008)Hansell, Bankier, MacMahon, McLoud, Muller, and Remy]{hansell2008fleischner}
David~M Hansell, Alexander~A Bankier, Heber MacMahon, Theresa~C McLoud, Nestor~L Muller, and Jacques Remy.
\newblock Fleischner society: glossary of terms for thoracic imaging.
\newblock \emph{Radiology}, 246\penalty0 (3):\penalty0 697--722, 2008.

\bibitem[Hu et~al.(2023)Hu, Liu, Zhao, Hou, Nie, and Li]{hu2023survey}
Linmei Hu, Zeyi Liu, Ziwang Zhao, Lei Hou, Liqiang Nie, and Juanzi Li.
\newblock A survey of knowledge enhanced pre-trained language models.
\newblock \emph{IEEE Transactions on Knowledge and Data Engineering}, 36\penalty0 (4):\penalty0 1413--1430, 2023.

\bibitem[Irvin et~al.(2019)Irvin, Rajpurkar, Ko, Yu, Ciurea-Ilcus, Chute, Marklund, Haghgoo, Ball, Shpanskaya, et~al.]{irvin2019chexpert}
Jeremy Irvin, Pranav Rajpurkar, Michael Ko, Yifan Yu, Silviana Ciurea-Ilcus, Chris Chute, Henrik Marklund, Behzad Haghgoo, Robyn Ball, Katie Shpanskaya, et~al.
\newblock Chexpert: A large chest radiograph dataset with uncertainty labels and expert comparison.
\newblock In \emph{Proceedings of the AAAI conference on artificial intelligence}, pages 590--597, 2019.

\bibitem[Johnson et~al.(2019)Johnson, Pollard, Greenbaum, Lungren, Deng, Peng, Lu, Mark, Berkowitz, and Horng]{johnson2019mimic}
Alistair~EW Johnson, Tom~J Pollard, Nathaniel~R Greenbaum, Matthew~P Lungren, Chih-ying Deng, Yifan Peng, Zhiyong Lu, Roger~G Mark, Seth~J Berkowitz, and Steven Horng.
\newblock {MIMIC-CXR-JPG}, a large publicly available database of labeled chest radiographs.
\newblock \emph{arXiv preprint arXiv:1901.07042}, 2019.

\bibitem[Kuang et~al.(2025)Kuang, Shen, Xie, Luo, Xu, Li, Li, Cheng, Lin, and Han]{kuang2025natural}
Jiayi Kuang, Ying Shen, Jingyou Xie, Haohao Luo, Zhe Xu, Ronghao Li, Yinghui Li, Xianfeng Cheng, Xika Lin, and Yu Han.
\newblock Natural language understanding and inference with mllm in visual question answering: A survey.
\newblock \emph{ACM Computing Surveys}, 57\penalty0 (8):\penalty0 1--36, 2025.

\bibitem[Li et~al.(2024{\natexlab{a}})Li, Zhang, Guo, Zhang, Li, Zhang, Zhang, Zhang, Li, Liu, et~al.]{li2024llava}
Bo Li, Yuanhan Zhang, Dong Guo, Renrui Zhang, Feng Li, Hao Zhang, Kaichen Zhang, Peiyuan Zhang, Yanwei Li, Ziwei Liu, et~al.
\newblock Llava-onevision: Easy visual task transfer.
\newblock \emph{arXiv preprint arXiv:2408.03326}, 2024{\natexlab{a}}.

\bibitem[Li et~al.(2023)Li, Wong, Zhang, Usuyama, Liu, Yang, Naumann, Poon, and Gao]{li2023llava}
Chunyuan Li, Cliff Wong, Sheng Zhang, Naoto Usuyama, Haotian Liu, Jianwei Yang, Tristan Naumann, Hoifung Poon, and Jianfeng Gao.
\newblock Llava-med: Training a large language-and-vision assistant for biomedicine in one day.
\newblock \emph{Advances in Neural Information Processing Systems}, 36:\penalty0 28541--28564, 2023.

\bibitem[Li et~al.(2022)Li, Li, Hu, and Tao]{li2022self}
Jun Li, Shibo Li, Ying Hu, and Huiren Tao.
\newblock A self-guided framework for radiology report generation.
\newblock In \emph{International Conference on Medical Image Computing and Computer-Assisted Intervention}, pages 588--598. Springer, 2022.

\bibitem[Li et~al.(2024{\natexlab{b}})Li, Su, Zhao, Lv, Wang, Navab, Hu, and Jiang]{li2024ultrasound}
Jun Li, Tongkun Su, Baoliang Zhao, Faqin Lv, Qiong Wang, Nassir Navab, Ying Hu, and Zhongliang Jiang.
\newblock Ultrasound report generation with cross-modality feature alignment via unsupervised guidance.
\newblock \emph{arXiv preprint arXiv:2406.00644}, 2024{\natexlab{b}}.

\bibitem[Li et~al.(2025)Li, Kim, M{\"u}ller, Felsner, Rueckert, Wiestler, Schnabel, and Bercea]{li2025language}
Jun Li, Su~Hwan Kim, Philip M{\"u}ller, Lina Felsner, Daniel Rueckert, Benedikt Wiestler, Julia~A Schnabel, and Cosmin~I Bercea.
\newblock Language models meet anomaly detection for better interpretability and generalizability.
\newblock In \emph{Medical Image Computing and Computer Assisted Intervention–MICCAI 2024 Workshops}, pages 1--11. Springer Nature Switzerland AG, 2025.

\bibitem[Lin et~al.(2023)Lin, Zhang, et~al.]{lin2023medical}
Zhihong Lin, Donghao Zhang, et~al.
\newblock Medical visual question answering: A survey.
\newblock \emph{Artificial Intelligence in Medicine}, 143:\penalty0 102611, 2023.

\bibitem[Liu et~al.(2023)Liu, Li, Wu, and Lee]{liu2023visual}
Haotian Liu, Chunyuan Li, Qingyang Wu, and Yong~Jae Lee.
\newblock Visual instruction tuning.
\newblock \emph{Advances in neural information processing systems}, 36:\penalty0 34892--34916, 2023.

\bibitem[Lu et~al.(2024)Lu, Liu, Zhang, Wang, Dong, Liu, Sun, Ren, Li, Yang, et~al.]{lu2024deepseek}
Haoyu Lu, Wen Liu, Bo Zhang, Bingxuan Wang, Kai Dong, Bo Liu, Jingxiang Sun, Tongzheng Ren, Zhuoshu Li, Hao Yang, et~al.
\newblock {DeepSeek-vl}: towards real-world vision-language understanding.
\newblock \emph{arXiv preprint arXiv:2403.05525}, 2024.

\bibitem[Lu and Wang(2025)]{LU2025103514}
Yinbin Lu and Alan Wang.
\newblock Integrating language into medical visual recognition and reasoning: A survey.
\newblock \emph{Medical Image Analysis}, 102:\penalty0 103514, 2025.

\bibitem[Luo et~al.(2024)Luo, Zhou, Royer, Sekuboyina, and Menze]{luo2024devide}
Haozhe Luo, Ziyu Zhou, Corentin Royer, Anjany Sekuboyina, and Bjoern Menze.
\newblock Devide: Faceted medical knowledge for improved medical vision-language pre-training.
\newblock \emph{arXiv preprint arXiv:2404.03618}, 2024.

\bibitem[Meissen et~al.(2023)Meissen, M{\"u}ller, Kaissis, et~al.]{meissen2023robust}
Felix Meissen, Philip M{\"u}ller, Georgios Kaissis, et~al.
\newblock Robust detection outcome: A metric for pathology detection in medical images.
\newblock In \emph{Medical Imaging with Deep Learning}, 2023.

\bibitem[Moor et~al.(2023)Moor, Huang, Wu, Yasunaga, Dalmia, Leskovec, Zakka, Reis, and Rajpurkar]{moor2023med}
Michael Moor, Qian Huang, Shirley Wu, Michihiro Yasunaga, Yash Dalmia, Jure Leskovec, Cyril Zakka, Eduardo~Pontes Reis, and Pranav Rajpurkar.
\newblock Med-flamingo: a multimodal medical few-shot learner.
\newblock In \emph{Machine Learning for Health (ML4H)}, pages 353--367. PMLR, 2023.

\bibitem[M{\"u}ller et~al.(2024)M{\"u}ller, Kaissis, and Rueckert]{muller2024chex}
Philip M{\"u}ller, Georgios Kaissis, and Daniel Rueckert.
\newblock Chex: Interactive localization and region description in chest {X-ray}.
\newblock In \emph{European Conference on Computer Vision}, pages 92--111. Springer, 2024.

\bibitem[Nguyen et~al.(2022)Nguyen, Lam, Le, Pham, Tran, Nguyen, Le, Pham, Tong, Dinh, et~al.]{nguyen2022vindr}
Ha~Q Nguyen, Khanh Lam, Linh~T Le, Hieu~H Pham, Dat~Q Tran, Dung~B Nguyen, Dung~D Le, Chi~M Pham, Hang~TT Tong, Diep~H Dinh, et~al.
\newblock Vindr-cxr: An open dataset of chest {X-rays} with radiologist’s annotations.
\newblock \emph{Scientific Data}, 9\penalty0 (1):\penalty0 429, 2022.

\bibitem[Padilla et~al.(2020)Padilla, Netto, and Da~Silva]{padilla2020survey}
Rafael Padilla, Sergio~L Netto, and Eduardo~AB Da~Silva.
\newblock A survey on performance metrics for object-detection algorithms.
\newblock In \emph{2020 International Conference on Systems, Signals and Image Processing (IWSSIP)}, pages 237--242. IEEE, 2020.

\bibitem[Radford et~al.(2021)Radford, Kim, Hallacy, Ramesh, Goh, Agarwal, Sastry, Askell, Mishkin, Clark, et~al.]{radford2021learning}
Alec Radford, Jong~Wook Kim, Chris Hallacy, Aditya Ramesh, Gabriel Goh, Sandhini Agarwal, Girish Sastry, Amanda Askell, Pamela Mishkin, Jack Clark, et~al.
\newblock Learning transferable visual models from natural language supervision.
\newblock In \emph{International conference on machine learning}, pages 8748--8763. PmLR, 2021.

\bibitem[Radiopaedia.org(2023)]{radiopaedia_radiomics}
Radiopaedia.org.
\newblock Radiomics.
\newblock \url{https://radiopaedia.org/}, 2023.

\bibitem[Sellergren et~al.(2025)Sellergren, Kazemzadeh, Jaroensri, Kiraly, Traverse, Kohlberger, Xu, Jamil, Hughes, Lau, et~al.]{sellergren2025medgemma}
Andrew Sellergren, Sahar Kazemzadeh, Tiam Jaroensri, Atilla Kiraly, Madeleine Traverse, Timo Kohlberger, Shawn Xu, Fayaz Jamil, C{\'\i}an Hughes, Charles Lau, et~al.
\newblock Medgemma technical report.
\newblock \emph{arXiv preprint arXiv:2507.05201}, 2025.

\bibitem[Shi et~al.(2024)Shi, Liu, Du, Wang, Wang, Guo, Ruan, Xu, Zhang, and Zhang]{shi-etal-2024-medical}
Xiaoming Shi, Zeming Liu, Li Du, Yuxuan Wang, Hongru Wang, Yuhang Guo, Tong Ruan, Jie Xu, Xiaofan Zhang, and Shaoting Zhang.
\newblock Medical dialogue system: A survey of categories, methods, evaluation and challenges.
\newblock In \emph{Findings of the Association for Computational Linguistics: ACL 2024}, pages 2840--2861, Bangkok, Thailand, 2024. Association for Computational Linguistics.

\bibitem[Shrestha et~al.(2023)Shrestha, Amgain, Khanal, Linte, and Bhattarai]{shrestha2023medical}
Prashant Shrestha, Sanskar Amgain, Bidur Khanal, Cristian~A Linte, and Binod Bhattarai.
\newblock Medical vision language pretraining: A survey.
\newblock \emph{arXiv preprint arXiv:2312.06224}, 2023.

\bibitem[Singhal et~al.(2025)Singhal, Tu, Gottweis, Sayres, Wulczyn, Amin, Hou, Clark, Pfohl, Cole-Lewis, et~al.]{singhal2025toward}
Karan Singhal, Tao Tu, Juraj Gottweis, Rory Sayres, Ellery Wulczyn, Mohamed Amin, Le Hou, Kevin Clark, Stephen~R Pfohl, Heather Cole-Lewis, et~al.
\newblock Toward expert-level medical question answering with large language models.
\newblock \emph{Nature Medicine}, pages 1--8, 2025.

\bibitem[Stefanini et~al.(2022)Stefanini, Cornia, et~al.]{stefanini2022show}
Matteo Stefanini, Marcella Cornia, et~al.
\newblock From show to tell: A survey on deep learning-based image captioning.
\newblock \emph{IEEE transactions on pattern analysis and machine intelligence}, 45\penalty0 (1):\penalty0 539--559, 2022.

\bibitem[Tanida et~al.(2023)Tanida, M{\"u}ller, Kaissis, and Rueckert]{tanida2023interactive}
Tim Tanida, Philip M{\"u}ller, Georgios Kaissis, and Daniel Rueckert.
\newblock Interactive and explainable region-guided radiology report generation.
\newblock In \emph{Proceedings of the IEEE/CVF Conference on Computer Vision and Pattern Recognition}, pages 7433--7442, 2023.

\bibitem[Team et~al.(2023)Team, Anil, Borgeaud, Alayrac, Yu, Soricut, Schalkwyk, Dai, Hauth, Millican, et~al.]{team2023gemini}
Gemini Team, Rohan Anil, Sebastian Borgeaud, Jean-Baptiste Alayrac, Jiahui Yu, Radu Soricut, Johan Schalkwyk, Andrew~M Dai, Anja Hauth, Katie Millican, et~al.
\newblock Gemini: a family of highly capable multimodal models.
\newblock \emph{arXiv preprint arXiv:2312.11805}, 2023.

\bibitem[Team et~al.(2025)Team, Kamath, Ferret, Pathak, Vieillard, Merhej, Perrin, Matejovicova, Ram{\'e}, Rivi{\`e}re, et~al.]{team2025gemma}
Gemma Team, Aishwarya Kamath, Johan Ferret, Shreya Pathak, Nino Vieillard, Ramona Merhej, Sarah Perrin, Tatiana Matejovicova, Alexandre Ram{\'e}, Morgane Rivi{\`e}re, et~al.
\newblock Gemma 3 technical report.
\newblock \emph{arXiv preprint arXiv:2503.19786}, 2025.

\bibitem[Wang et~al.(2022)Wang, Zhou, Wang, Vardhanabhuti, and Yu]{wang2022multi}
Fuying Wang, Yuyin Zhou, Shujun Wang, Varut Vardhanabhuti, and Lequan Yu.
\newblock Multi-granularity cross-modal alignment for generalized medical visual representation learning.
\newblock \emph{Advances in neural information processing systems}, 35:\penalty0 33536--33549, 2022.

\bibitem[Wang et~al.(2024)Wang, Bai, Tan, Wang, Fan, Bai, Chen, Liu, Wang, Ge, et~al.]{wang2024qwen2}
Peng Wang, Shuai Bai, Sinan Tan, Shijie Wang, Zhihao Fan, Jinze Bai, Keqin Chen, Xuejing Liu, Jialin Wang, Wenbin Ge, et~al.
\newblock Qwen2-vl: Enhancing vision-language model's perception of the world at any resolution.
\newblock \emph{arXiv preprint arXiv:2409.12191}, 2024.

\bibitem[Wu et~al.(2023)Wu, Zhang, Zhang, Wang, and Xie]{wu2023medklip}
Chaoyi Wu, Xiaoman Zhang, Ya Zhang, Yanfeng Wang, and Weidi Xie.
\newblock Medklip: Medical knowledge enhanced language-image pre-training for x-ray diagnosis.
\newblock In \emph{Proceedings of the IEEE/CVF international conference on computer vision}, pages 21372--21383, 2023.

\bibitem[Wu et~al.(2024)Wu, Wu, Zheng, and Yang]{wu2024medkp}
Jiageng Wu, Xian Wu, Yefeng Zheng, and Jie Yang.
\newblock Medkp: Medical dialogue with knowledge enhancement and clinical pathway encoding.
\newblock \emph{arXiv preprint arXiv:2403.06611}, 2024.

\bibitem[Wu et~al.(2021)Wu, Agu, Lourentzou, Sharma, Paguio, Yao, Dee, Mitchell, Kashyap, Giovannini, et~al.]{wu2021chest}
Joy~T Wu, Nkechinyere~N Agu, Ismini Lourentzou, Arjun Sharma, Joseph~A Paguio, Jasper~S Yao, Edward~C Dee, William Mitchell, Satyananda Kashyap, Andrea Giovannini, et~al.
\newblock Chest imagenome dataset for clinical reasoning.
\newblock \emph{arXiv preprint arXiv:2108.00316}, 2021.

\bibitem[Xiao et~al.(2024{\natexlab{a}})Xiao, Wu, Xu, et~al.]{xiao2024florence}
Bin Xiao, Haiping Wu, Weijian Xu, et~al.
\newblock Florence-2: Advancing a unified representation for a variety of vision tasks.
\newblock In \emph{Proceedings of the IEEE/CVF Conference on Computer Vision and Pattern Recognition}, pages 4818--4829, 2024{\natexlab{a}}.

\bibitem[Xiao et~al.(2024{\natexlab{b}})Xiao, Yang, Lan, et~al.]{xiao2024towards}
Linhui Xiao, Xiaoshan Yang, Xiangyuan Lan, et~al.
\newblock Towards visual grounding: A survey.
\newblock \emph{arXiv preprint arXiv:2412.20206}, 2024{\natexlab{b}}.

\bibitem[Ye et~al.(2024)Ye, Wang, Li, Deng, Li, Li, Duan, Huang, Su, Wang, et~al.]{ye2024gmai}
Jin Ye, Guoan Wang, Yanjun Li, Zhongying Deng, Wei Li, Tianbin Li, Haodong Duan, Ziyan Huang, Yanzhou Su, Benyou Wang, et~al.
\newblock Gmai-mmbench: A comprehensive multimodal evaluation benchmark towards general medical ai.
\newblock \emph{Advances in Neural Information Processing Systems}, 37:\penalty0 94327--94427, 2024.

\bibitem[Zhang et~al.(2023{\natexlab{a}})Zhang, Wu, Zhang, Xie, and Wang]{zhang2023knowledge}
Xiaoman Zhang, Chaoyi Wu, Ya Zhang, Weidi Xie, and Yanfeng Wang.
\newblock Knowledge-enhanced visual-language pre-training on chest radiology images.
\newblock \emph{Nature Communications}, 14\penalty0 (1):\penalty0 4542, 2023{\natexlab{a}}.

\bibitem[Zhang et~al.(2023{\natexlab{b}})Zhang, Wu, Zhao, et~al.]{zhang2023pmc}
Xiaoman Zhang, Chaoyi Wu, Ziheng Zhao, et~al.
\newblock {Pmc-VQA}: Visual instruction tuning for medical visual question answering.
\newblock \emph{arXiv preprint arXiv:2305.10415}, 2023{\natexlab{b}}.

\bibitem[Zhu et~al.(2025)Zhu, Wang, Chen, Liu, Ye, Gu, Tian, Duan, Su, Shao, et~al.]{zhu2025internvl3}
Jinguo Zhu, Weiyun Wang, Zhe Chen, Zhaoyang Liu, Shenglong Ye, Lixin Gu, Hao Tian, Yuchen Duan, Weijie Su, Jie Shao, et~al.
\newblock Internvl3: Exploring advanced training and test-time recipes for open-source multimodal models.
\newblock \emph{arXiv preprint arXiv:2504.10479}, 2025.

\end{thebibliography}
}
\clearpage
\newpage
% \documentclass[10pt,twocolumn,letterpaper]{article}
% \usepackage[rebuttal,applications]{wacv}  % use this for an Applications Track rebuttal
%\usepackage[rebuttal,algorithms]{wacv}  % use this for an Algorithms Track rebuttal

% % Include other packages here, before hyperref.
% \usepackage{graphicx}
% \usepackage{amsmath}
% \usepackage{amssymb}
% \usepackage{booktabs}
% \usepackage{mathrsfs}
% \usepackage[ruled,linesnumbered]{algorithm2e}
% \usepackage{mathptmx} 
% \usepackage{bookmark} 
% \usepackage{adjustbox}

% \usepackage[most]{tcolorbox}  % enables advanced tcolorbox features
% \definecolor{lightblue}{RGB}{16,98,180}
% \definecolor{lightpink}{RGB}{243,40,109}
% \hypersetup{
%   colorlinks=true,
%   linkcolor=lightpink,
%   citecolor=lightblue,
%   urlcolor=lightpink,
%   filecolor=lightblue,
%   pdftitle={Your Paper Title},
%   pdfauthor={Your Name},
%   pdfstartview=FitH,
%   bookmarksdepth=4
% }

% Import additional packages in the preamble file, before hyperref
% \input{preamble}

% If you comment hyperref and then uncomment it, you should delete
% egpaper.aux before re-running latex.  (Or just hit 'q' on the first latex
% run, let it finish, and you should be clear).

\lstdefinestyle{jsonstyle}{
  basicstyle=\ttfamily\small,
  breaklines=true,
  frame=none,
  backgroundcolor=\color{gray!5},
  literate=
   *{0}{{{\color{black}0}}}{1}
    {1}{{{\color{black}1}}}{1}
    {2}{{{\color{black}2}}}{1}
    {3}{{{\color{black}3}}}{1}
    {4}{{{\color{black}4}}}{1}
    {5}{{{\color{black}5}}}{1}
    {6}{{{\color{black}6}}}{1}
    {7}{{{\color{black}7}}}{1}
    {8}{{{\color{black}8}}}{1}
    {9}{{{\color{black}9}}}{1}
    {:}{{{\color{black}:}}}{1}
    {,}{{{\color{black},}}}{1}
    {"}{{{\color{black}"}}}{1}
}

% If you wish to avoid re-using figure, table, and equation numbers from
% the main paper, please uncomment the following and change the numbers
% appropriately.
%\setcounter{figure}{2}
%\setcounter{table}{1}
%\setcounter{equation}{2}
% If you wish to avoid re-using reference numbers from the main paper,
% please uncomment the following and change the counter value to the
% number of references you have in the main paper (here, 100).
%\makeatletter
%\apptocmd{\thebibliography}{\global\c@NAT@ctr 100\relax}{}{}
%\makeatother

% %%%%%%%%% PAPER ID  - PLEASE UPDATE
% \def\wacvPaperID{*****} % *** Enter the WACV Paper ID here
% \def\confName{WACV}
% \def\confYear{2026}

\tcbset{
  k2boxstyle/.style={
    colback=cyan!10!white,   
    colframe=cyan!60!black, 
    title=#1,
    rounded corners,        
    enhanced,
    box align=base,
    halign=flush left,
    left=2pt,
    right=2pt,
    top=2pt,
    bottom=2pt
  }
}

% \begin{document}

%%%%%%%%% TITLE - PLEASE UPDATE
% \title{\LaTeX\ Guidelines for Author Response}  % **** Enter the paper title here
% \title{Knowledge to Sight: Reasoning over Visual Attributes through Knowledge \\
% Decomposition for Abnormality Grounding \\
% \vspace{0.3em} {\textit{Supplementary Material}}
% }

\twocolumn[
  \begin{center}
    \textbf{\large Knowledge to Sight: Reasoning over Visual Attributes through Knowledge\\[0.3em]
    Decomposition for Abnormality Grounding} \\
     \vspace{1.2em} {\large {Supplementary Material}}
  \end{center}
  \vspace{1em}
]

% \maketitle
% \thispagestyle{empty}
% \appendix

%%%%%%%%% BODY TEXT - ENTER YOUR RESPONSE BELOW
\tableofcontents

%-------------------------------------------------------------------------

\section{Dataset Details}

In this section, we provide detailed statistics and data characteristics of the two datasets used in our experiments: VinDr-CXRR~\cite{nguyen2022vindr} and PadChest-GR~\cite{castro2024padchest}. This supplement presents additional information relevant to our evaluation.

\subsection{Details of VinDr-CXR}
\begin{figure}[ht]
  \centering
  \includegraphics[width=\linewidth]{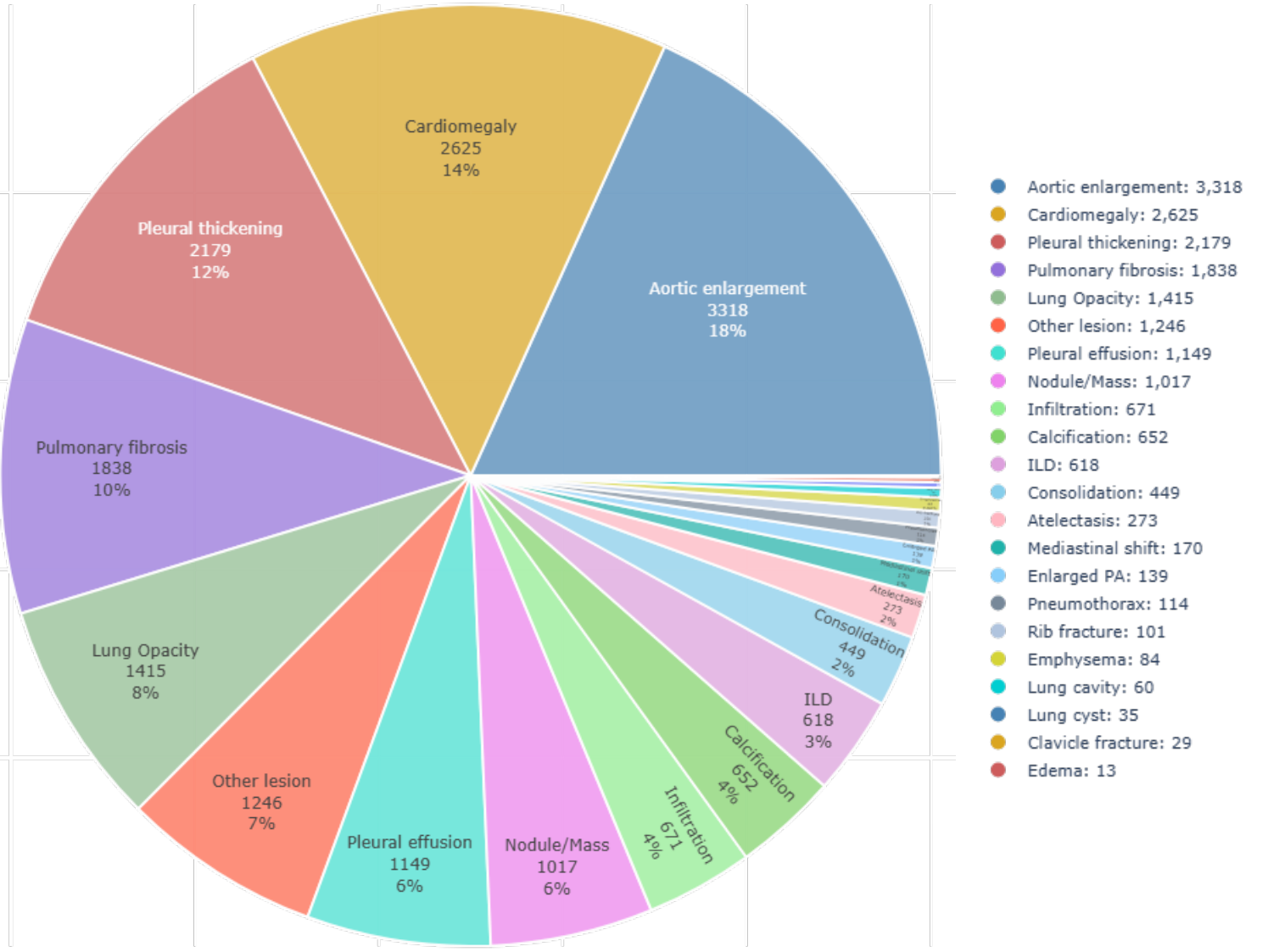}
  \caption{Distribution of the 22 abnormality types in the VinDr-CXR dataset. The dataset exhibits a long-tailed distribution. The dataset is split into a training set of 16,087 samples and a held-out test set of 2,108 samples.}
  \label{fig:distribution_vindr}
\end{figure}

We chose VinDr-CXR as the training source because it is annotated by experienced radiologists. We use all available image–abnormality pairs, resulting in a total of 18,195 samples covering 22 distinct thoracic findings. Each image may contain one or more annotated abnormalities.
To address annotation inconsistencies across multiple radiologists, we adopt the weighted box fusion strategy~\cite{muller2024chex}, following prior work~\cite{deperrois2025radvlm,bannur2024maira}, to merge overlapping bounding boxes into a unified set per abnormality per image. We follow the official split provided by the dataset, resulting in a training set of 16,087 samples and a test set of 2,108 samples.
Figure~\ref{fig:distribution_vindr} shows the distribution of the dataset in our experiments.

\subsection{Details of PadChest-GR}

\begin{figure}[ht]
  \centering
  \includegraphics[width=\linewidth]{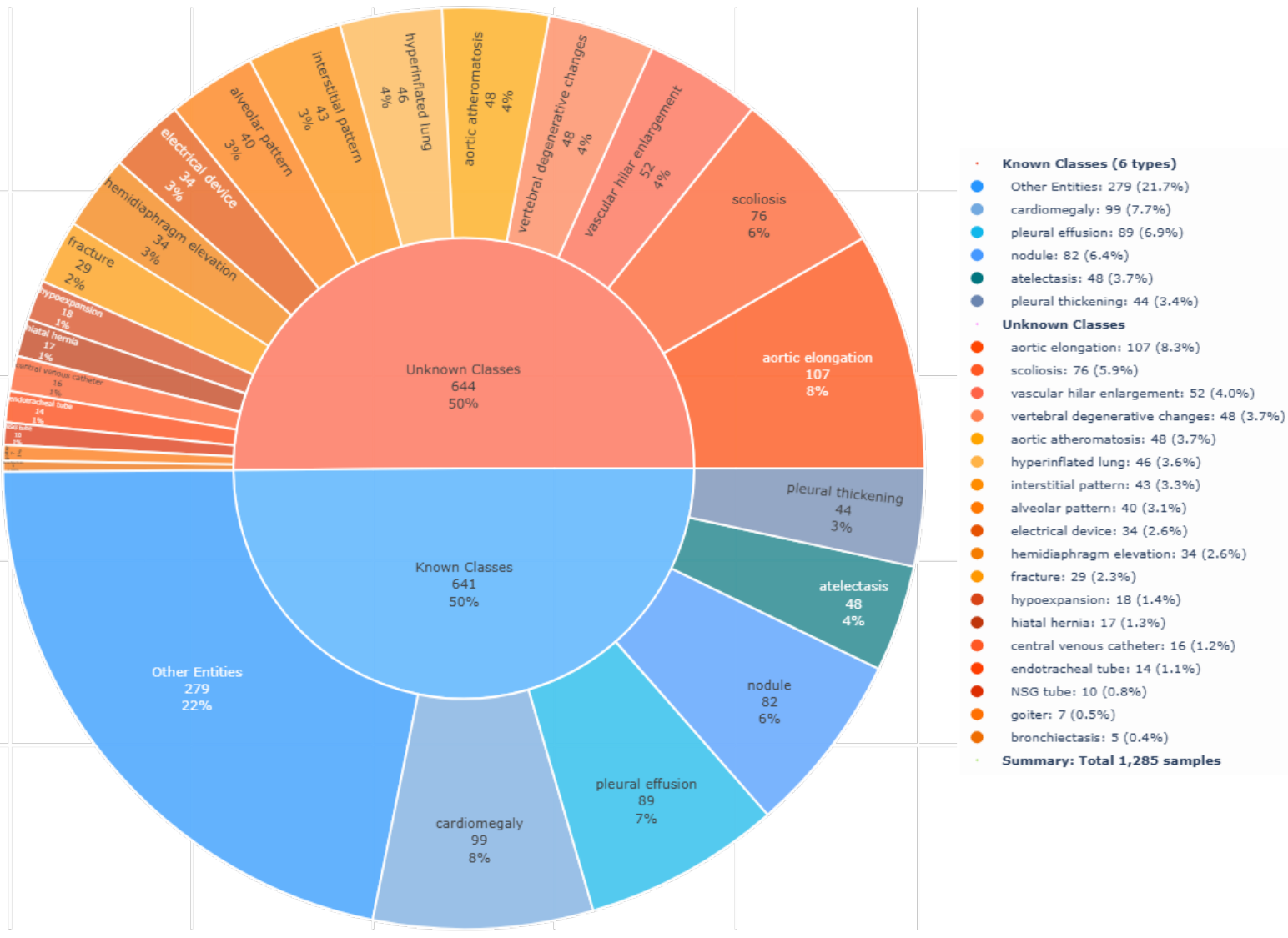}
  \caption{Distribution of the 24 abnormality types in the PadChest-GR dataset. Known classes are mapped to VinDr-CXR labels, while the rest are treated as unknown for OOD evaluation.}
  \label{fig:distribution_padchest}
\end{figure}

PadChest-GR~\cite{castro2024padchest} is used exclusively for evaluation. We construct two evaluation subsets from its official test split (covering 24 abnormality types): one for standard zero-shot generalization and another for out-of-distribution (OOD) evaluation.
The zero-shot subset contains 641 image–abnormality pairs spanning 6 types, labeled with \textbf{known} categories that are also present in VinDr-CXR. The OOD subset includes 644 image–abnormality pairs across 18 types labeled with \textbf{unknown} categories, which are not seen during training. This split enables controlled evaluation of both generalization within known classes and robustness to novel, unseen concepts.
Based on semantic and clinical correspondence with VinDr-CXR, we categorize 6 PadChest-GR labels as known classes (e.g., \textit{cardiomegaly}, \textit{atelectasis}, \textit{nodule}), while the remaining 18 (e.g., \textit{scoliosis}, \textit{aortic atheromatosis}, \textit{electrical device}) are treated as unknown. 
Figure~\ref{fig:distribution_padchest} illustrates the distribution of the PadChest-GR categories, highlighting the long-tailed nature of both known and unknown subsets.

\subsection{Annotation Preprocess for Florence-2}

We follow the official formatting protocol of Florence-2 for converting bounding box annotations into language-based location prompts and answers.

Each bounding box is represented in the standard format as pixel coordinates $(x_1, y_1, x_2, y_2)$ on an image of width $W$ and height $H$. These coordinates are first normalized to the range $[0, 1000]$ using the following transformation:
\begin{equation}
    \texttt{loc\_x} = \left\lfloor \frac{x}{W} \times 1000 \right\rfloor, \quad
    \texttt{loc\_y} = \left\lfloor \frac{y}{H} \times 1000 \right\rfloor,
\end{equation}
where $\lfloor \cdot \rfloor$ denotes the floor operation to obtain discrete integer values. The normalized bounding box is then represented as a sequence of location tokens:
\begin{equation}
    \texttt{Label} \ \langle \texttt{loc\_x1} \rangle \langle \texttt{loc\_y1} \rangle \langle \texttt{loc\_x2} \rangle \langle \texttt{loc\_y2} \rangle
\end{equation}

During training and evaluation, each image–annotation pair is formatted into a prompt–answer style. The prompt queries the presence and location of abnormalities, and the answer consists of one or more spatial tokens corresponding to the bounding boxes.

\begin{tcolorbox}[k2boxstyle={Florence-2 (base) pair}]
\textbf{Prompt:} Locate disease \{Disease\}. \\
\textbf{Answer:} \\
Disease $\langle$loc\_145$\rangle$ $\langle$loc\_300$\rangle$ $\langle$loc\_812$\rangle$ $\langle$loc\_940$\rangle$ \\
Disease $\langle$loc\_201$\rangle$ $\langle$loc\_322$\rangle$ $\langle$loc\_715$\rangle$ $\langle$loc\_850$\rangle$
\end{tcolorbox}

In our K2Sight framework, we further decompose the underlying knowledge associated with each disease entity and extract visual-oriented descriptions. These include concise phrases describing shape, texture, and anatomical location for each abnormality. The final prompts are augmented with these definitions to enhance model during training. The training pairs follow the format below:

\begin{tcolorbox}[k2boxstyle={K2Sight-Lite Pair}]
\textbf{Prompt:} Locate disease \{Disease\}, which means \{attribute-based description\}. \\
\textbf{Answer:} \\
Disease $\langle$loc\_145$\rangle$ $\langle$loc\_300$\rangle$ $\langle$loc\_812$\rangle$ $\langle$loc\_940$\rangle$ \\
Disease $\langle$loc\_201$\rangle$ $\langle$loc\_322$\rangle$ $\langle$loc\_715$\rangle$ $\langle$loc\_850$\rangle$
\end{tcolorbox}

\subsection{Annotation Preprocess for Qwen2-VL-Instruct}

We follow the official guideline of Qwen2-VL-2B-Instruct for converting bounding boxes into text pairs. Similar to Florence-2, each bounding box is expressed as $(x_1, y_1, x_2, y_2)$ and normalized by dividing each coordinate by the image width or height and multiplying by 1000:
\begin{equation}
    \texttt{new\_x} = \left\lfloor \frac{x}{W} \times 1000 \right\rfloor, \quad
    \texttt{new\_y} = \left\lfloor \frac{y}{H} \times 1000 \right\rfloor.
\end{equation}

Unlike Florence-2, Qwen2-VL-Instruct directly utilizes 2D bounding box coordinates in the output and does not require special format. The prompt structure is derived from the official Qwen2-VL-Instruct repository, where bounding box annotations are requested in JSON format.

\begin{tcolorbox}[k2boxstyle={Qwen2-VL (base) Pair}]
\textbf{Prompt:} \\
Return bounding boxes of 'Disease' areas as JSON format:
\begin{lstlisting}[style=jsonstyle]
[{"bbox_2d": [x1, y1, x2, y2], "label": "label"}, ...]
\end{lstlisting}
\textbf{Answer:}
\begin{lstlisting}[style=jsonstyle]
[{"bbox_2d": [276, 141, 484, 218], "label": "Disease"}, 
{"bbox_2d": [552, 127, 767, 230], "label": "Disease"}]
\end{lstlisting}
\end{tcolorbox}

In our K2Sight, we enrich the original prompt–answer structure by appending the attribute-based description at the end. The format of the training pairs is shown below:

\begin{tcolorbox}[k2boxstyle={K2Sight-Base Pair}]
\textbf{Prompt:} \\
Return bounding boxes of 'Disease' areas as JSON format:
\begin{lstlisting}[style=jsonstyle]
[{"bbox_2d": [x1, y1, x2, y2], "label": "label"}, ...]
\end{lstlisting}
Note: \{attribute-based description\} \\[4pt]
\textbf{Answer:}
\begin{lstlisting}[style=jsonstyle]
[{"bbox_2d": [276, 141, 484, 218], "label": "Disease"},
{"bbox_2d": [552, 127, 767, 230], "label": "Disease"}]
\end{lstlisting}
\end{tcolorbox}

%------------------------------------------------------------------------
\section{Further Implementation Details of K2Sight}

\subsection{K2Sight Framework Pseudocode}
We further outline the complete pseudocode of the K2Sight framework. It consists of two core stages. In the first stage, the Knowledge Decomposition Constructor extracts and distills clinical definitions into prompts that are visually informative. The second stage, Semantic-Guided Fine-Tuning, fine-tunes vision-language models using these prompts to improve localization accuracy.

\begin{algorithm}[htbp]
\caption{K2Sight Framework}
\label{alg:k2sight}
\KwIn{Abnormality definitions $\{d(a)\}_{a \in {A}}$; training dataset ${D}_{\text{train}} = \{(I_i, B_i)\}$}
\KwOut{Our K2Sight model ${M}$}
\vspace{2pt}

\textbf{Stage 1: Knowledge Decomposition Constructor} \\
\ForEach{$a \in {A}$}{
    Retrieve textual definition $d(a)$ from medical sources to get the medical definition\;
    Generate structured prompt $\pi(a, d(a))$ incorporating visual attributes (shape, intensity, density, location)\;
    Sample $N=5$ candidate visual descriptions $\{\tilde{k}_i(a)\}_{i=1}^N$ using GPT-4o with controlled decoding\;
    Select best candidate $k(a) \leftarrow \text{HumanSelect}(\{\tilde{k}_i(a)\})$\;
}
Construct prompt dictionary ${K} = \{a \mapsto k(a)\}$\;

\vspace{4pt}
\textbf{Stage 2: Semantic-Guided Fine-Tuning} \\
\ForEach{$(I_i, B_i) \in {D}_{\text{train}}$}{
    \ForEach{abnormality $a$ in $B_i$}{
        Format instruction using $k(a)$ and convert bounding boxes to token sequence $Y$\;
        Train ${M}$ to predict $Y$ given image $I_i$ and prompt $k(a)$ via cross-entropy loss\;
    }
}
\Return{${M}$}
\end{algorithm}

\subsection{Clinical Definition Collection}
We collect textual definitions for each abnormality class from authoritative radiology resources. Specifically, we extract formal descriptions from the official documentation of VinDr-CXR~\cite{nguyen2022vindr} and publicly available entries on Radiopaedia~\cite{radiopaedia_radiomics}, a widely used collaborative radiology reference. All definitions are used under the Creative Commons BY-NC-SA 3.0 license for research purposes. Each definition is manually verified to ensure medical accuracy, visual interpretability, and alignment with radiographic appearance.

\subsection{Visual Attribute Extraction Parameters}

We use GPT-4o for visual attribute extraction in a zero-shot setting. For each abnormality definition, we generate $N=5$ candidate descriptions per class using the following decoding configuration: temperature = 0.7, top-p = 0.7, repetition penalty = 1.1, and a maximum output length of 1024 tokens. Each generation is conditioned on a structured prompt template (see Sec.~3.1 in the main paper), which encourages the model to focus on four core visual aspects: shape, intensity, appearance pattern, and anatomical location.
After generation, minimal text cleaning is applied to remove formatting artifacts or incomplete fragments. Each candidate set is then reviewed by a domain-aware annotator, who selects the most visually faithful, semantically aligned, and concise description. Candidates that are speculative, overly verbose, or dependent on latent clinical information are discarded.
The final attribute-grounded prompts are consistently used across training and inference for all models in the K2Sight framework. Full lists of the selected descriptions for VinDr-CXR and PadChest-GR abnormality classes are provided in Table~\ref{tab:attribute_vindr} and Table~\ref{tab:attribute_padchest}, respectively.

\subsection{Training Details}

\noindent \textbf{Florence-2 Base and K2Sight Lite.} Both models are trained on the VinDr-CXR training set, which contains 16,087 annotated image–abnormality pairs. To ensure a controlled comparison and to isolate the effect of knowledge-enhanced instructions, we adopt identical training configurations for both models. The fine-tuning process spans 20 epochs, using the AdamW optimizer with an initial learning rate of $3 \times 10^{-6}$ and a weight decay coefficient of 0.01 to prevent overfitting. Training is executed on two A100 GPUs with a batch size of 16 per device (resulting in an effective batch size of 32), and distributed data parallelism is enabled via PyTorch Lightning. We employ mixed-precision (FP16) training for compute efficiency. \\

\noindent \textbf{Qwen2-VL 3B and K2Sight Base.}
The Qwen2-VL (base) and our K2Sight Base model are both fully fine-tuned, meaning that the vision encoder, multimodal projector, and language decoder are jointly updated during training. Both variants use the same training configuration to ensure direct comparability; the only difference lies in the input format, in which K2Sight uses structured prompts incorporating attribute-based descriptions, while the baseline version uses only raw class names. Fine-tuning is conducted using DeepSpeed ZeRO-3 optimization on 4×A100 GPUs to enable efficient large-scale training with a reduced memory footprint. Each model is trained for 20 epochs, with a per-device batch size of 8 and a gradient accumulation step size of 8, yielding a total effective batch size of 256. The learning rate is initialized at $3 \times 10^{-5}$ and follows a cosine decay schedule with a warmup ratio of 0.1. All training runs use bfloat16 precision to accelerate computation.

\subsection{Inference Settings}
To ensure a standardized and unbiased evaluation, we adopt a uniform decoding configuration across all model variants. In particular, we use greedy decoding with the temperature fixed at 0.0, disabling all stochastic sampling strategies such as top-k sampling. This deterministic decoding approach ensures consistency in output across runs and models. The maximum output length is set to 1024 tokens for both Florence-2 and Qwen2-VL variants, which is sufficient to accommodate all predicted bounding-box sequences. 
All inference experiments are conducted on a single A100 GPU per model to ensure resource parity. During evaluation, we use a batch size of 32, which provides a good trade-off between speed and memory usage while maintaining consistent throughput across all models.

%------------------------------------------------------------------------
\section{Experiments}
\subsection{Comparison Model Checkpoints}

All comparision models are using their publicly available checkpoints. These include general-purpose models such as Qwen2-VL~\cite{wang2024qwen2} and InternVL3~\cite{zhu2025internvl3}, as well as domain-specific baselines like RadVLM~\cite{deperrois2025radvlm} and MAIRA-2~\cite{bannur2024maira}. All evaluations are conducted in a zero-shot or fine-tuned setting as described in the main experiments.

\begin{itemize}
  \item \textbf{Qwen2-VL-2B-Instruct}: \url{https://huggingface.co/Qwen/Qwen2-VL-2B-Instruct}
  \item \textbf{Qwen2-VL-7B-Instruct}: \url{https://huggingface.co/Qwen/Qwen2-VL-7B-Instruct}
  \item \textbf{InternVL3-2B}: \url{https://huggingface.co/OpenGVLab/InternVL3-2B}
 
  \item \textbf{RadVLM (7B)}: \url{https://huggingface.co/KrauthammerLab/RadVLM}
  \item \textbf{MAIRA-2 (13B)}: \url{https://huggingface.co/microsoft/maira-2}
\end{itemize}

\subsection{Attribute-Conditioned Ablation Study}

For the ablation study described in Section~4.2 of the main paper, we perform attribute-conditioned prompt masking to investigate the individual contribution of four canonical visual attributes: shape, density, intensity, and location. Specifically, we construct curated lexicons for each attribute category and remove all matched terms from the structured prompts at inference time. This simulates the absence of a specific attribute while keeping the rest of the prompt intact. For detailed lists of attribute-specific terms used in the masking procedure, please refer to Table~\ref{tab:attribute_terms}.

\subsection{Further Results on Generalist VLMs with Our Enhanced Prompts}

In this section, we further evaluate the performance of generalist VLMs using our enhanced visual-oriented prompts, distilled from medical knowledge via our K2Sight framework. Table~\ref{tab:further_general_vlm_performance} presents the evaluation results. Compared to directly using zero-shot prompts with only disease entity names, the performance improves slightly in some cases. For example, Qwen2-VL-7B improves from 1.52 to 2.27 in $mAP_{30}$ and from 0.48 to 1.07 in $mAP_{50}$ on VinDR-CXR dataset.
However, some models show a performance drop. For instance, InternVL3-8B decreases from 0.24 to 0.18 in $mAP_{30}$, and from 0.04 to 0.03 in $mAP_{50}$.

Overall, relying solely on prompt engineering without further incorporating such descriptive prompts into semantically guided training is still insufficient to elevate generalist models to the performance level of medical specialists.

\begin{table}[t]
\renewcommand{\arraystretch}{1.1}
  \centering
  \caption{Further evaluation for generalist models with our visual-oriented enhanced prompt. All settings are the same as in the main paper Sec. 4.1.}
  \begin{adjustbox}{max width=\linewidth}
  \begin{tabular}{l | ccc | ccc}
    \toprule
    \textbf{Model} & \multicolumn{3}{c|}{VinDr-CXR} & \multicolumn{3}{c}{PadChest-GR} \\
    & mAP$_{30}$ & mAP$_{50}$ & mAP$_{75}$ & mAP$_{30}$ & mAP$_{50}$ & mAP$_{75}$ \\
    \midrule
    Qwen2-VL-7B & \textbf{2.27} & \textbf{1.07} & \textbf{0.02} & \textbf{1.91} & \textbf{0.26} & \textbf{0.06} \\
    Qwen2-VL-2B & 0.19 & 0.04 & 0.01 & 0.25 & 0.08 & 0.00 \\
    \midrule
    InternVL3-8B & 0.18 & 0.03 & 0.00 & 0.64 & 0.24 & 0.00 \\
    InternVL3-2B & 0.15 & 0.02 & 0.00 & 0.47 & 0.12 & 0.00 \\
    \bottomrule
  \end{tabular}
  \end{adjustbox}
  \label{tab:further_general_vlm_performance}
\end{table}

\subsection{More Visualization Results}

\begin{figure*}[htbp]
  \centering
  \includegraphics[width=\linewidth]{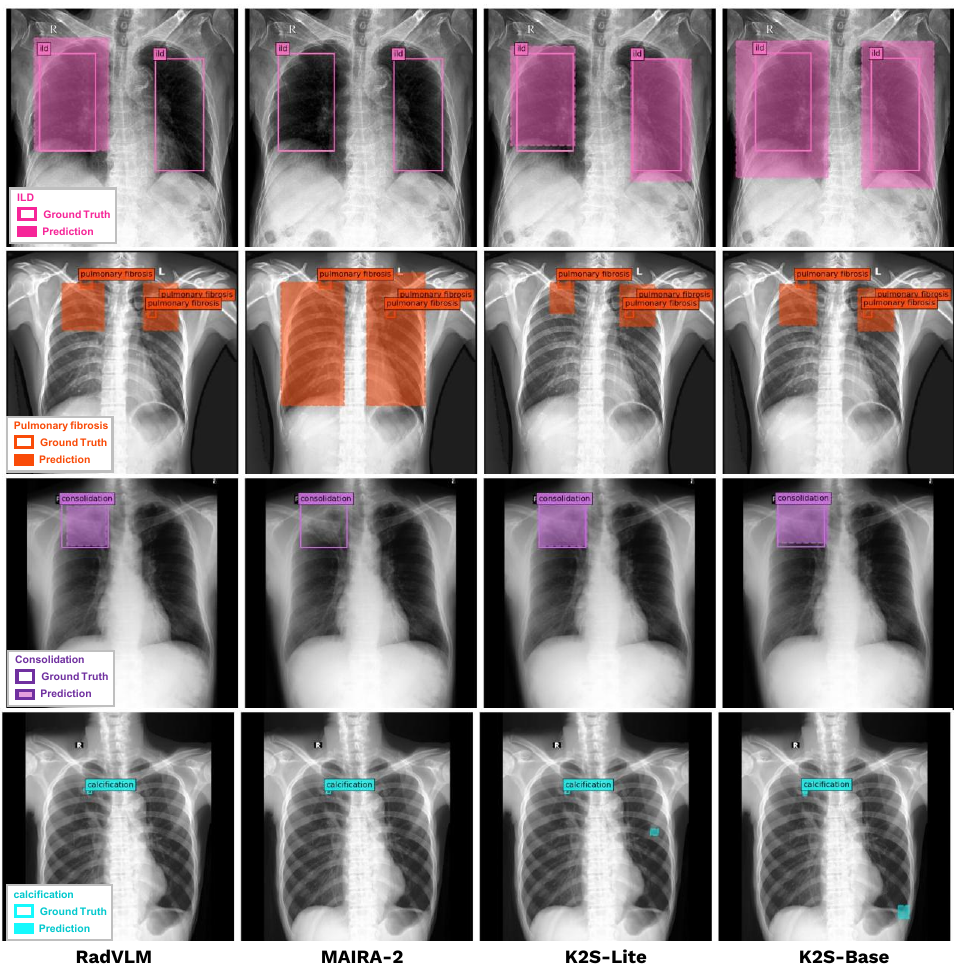}
  \caption{
  Qualitative examples of abnormality grounding across models.}
  \label{fig:more_examples1}
\end{figure*}

\begin{figure*}[ht]
  \centering
  \includegraphics[width=\linewidth]{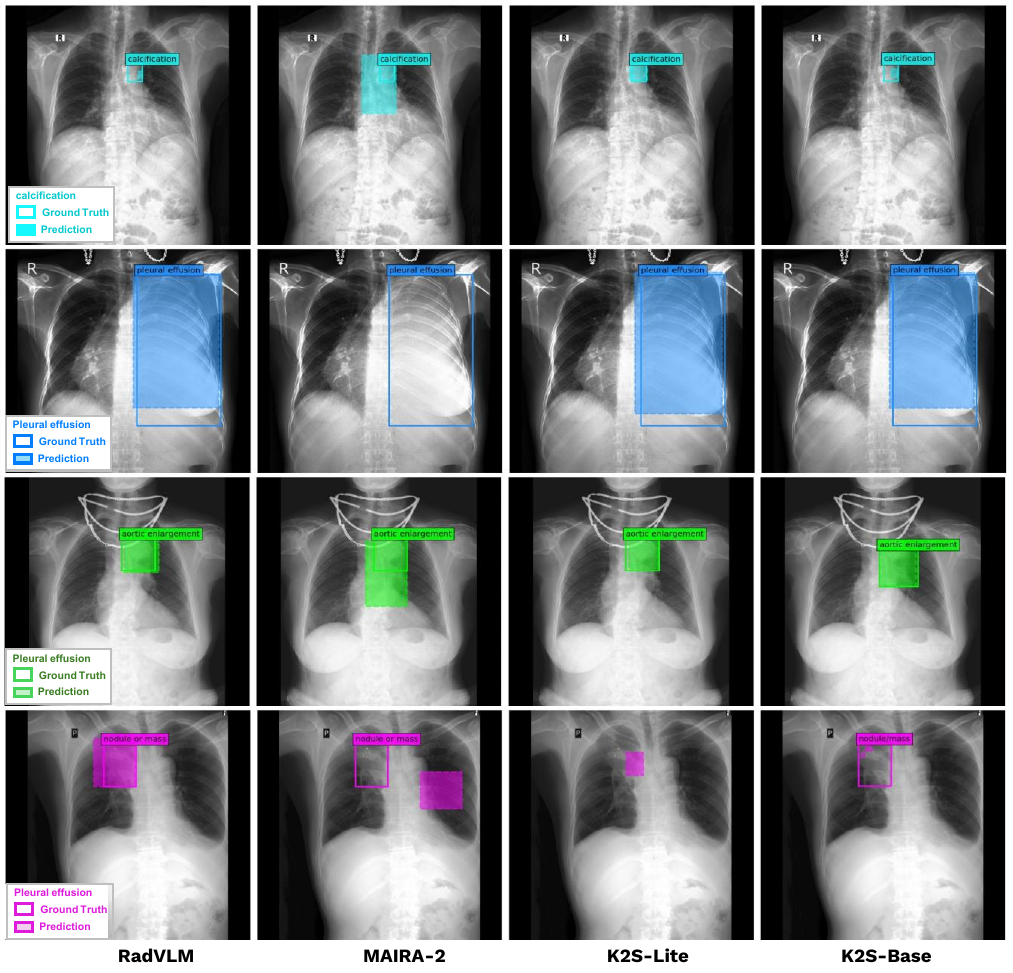}
  \caption{
  Qualitative examples of abnormality grounding across models.}
  \label{fig:more_examples2}
\end{figure*}

We provide additional qualitative results comparing K2Sight variants with MAIRA-2 and RadVLM. As shown in Figure~\ref{fig:more_examples1} and Figure~\ref{fig:more_examples2}, K2Sight consistently produces more accurate and compact localizations, with higher alignment to radiologist-annotated regions. 
These examples further illustrate the benefit of attribute-guided prompts in improving grounding fidelity, especially for subtle or structurally complex abnormalities.

%--------------------------------Table and figure ----------------------------------

\label{tab:attribute_terms}
\begin{table}[ht]
\centering
\caption{Term sets used to mask individual visual attributes during inference-time ablation.}
\renewcommand{\arraystretch}{1.2}
\begin{tabular}{@{}p{2cm} p{6cm}@{}}
\toprule
\textbf{Attribute} & \textbf{Terms Used for Masking} \\
\toprule
\textbf{Shape} & 
round, oval, circular, spherical, elliptical, triangular, rectangular, linear, curved, straight, irregular, lobulated, spiculated, nodular, stellate, mass-like, lump-like, reticular, honeycomb, septal, branching, wedge-shaped, crescentic, patchy, diffuse, borders, contour, outline, edge, pattern, irregularity \\
\midrule
\textbf{Density} & 
dense, solid, soft-tissue, fluid, liquid, gas, air-filled, air-containing, fat-density, calcified, calcific, ossified, consolidated, radiopaque, radiolucent, sclerotic, fibrotic, thick, thin, firm, density \\
\midrule
\textbf{Intensity} & 
bright, white, hyperdense, hyperintense, high-signal, dark, black, hypodense, hypointense, low-signal, gray, greyish, hazy, faint, subtle, opaque, lucent, transparent, prominent, clear, ground-glass, increased, decreased, reduced, diminished \\
\midrule
\textbf{Location} & 
within the lung, pleural cavity, pleural space, pulmonary artery, lung tissue, lung fields, in the lung, supradiaphragmatic, intrathoracic, extrathoracic, paramediastinal, paravertebral, mediastinum, mediastinal, costophrenic, retrocardiac, peripheral, perihilar, subpleural, unilateral, bilateral, central, aorta, heart, basal, posterior, anterior, ventral, dorsal, apical, middle, lower, upper, medial, lateral, right, left \\
\bottomrule
\end{tabular}

\end{table}

\begin{table*}[htbp]
\centering
\caption{Final attribute-based visual descriptions for each abnormality in VinDr-CXR.}
\label{tab:attribute_vindr}
\renewcommand{\arraystretch}{1.3}
\begin{tabular}{p{5cm} p{10.5cm}}
\toprule
\textbf{Abnormality} & \textbf{Attribute-based Visual Description} \\
\midrule
Aortic Enlargement & Widening of the aorta visible as an enlarged artery on imaging. \\
Atelectasis & Collapsed lung tissue causing darkened or shrunken areas in the lung. \\
Cardiomegaly & Enlargement of the heart seen when the heart appears larger than normal. \\
Calcification & Calcium deposits in lung tissue visible as bright white spots. \\
Clavicle Fracture & A break in the collarbone seen as a gap or irregularity in the bone. \\
Consolidation & Lung tissue filled with fluid or cells causing dense solid areas on imaging. \\
Edema & Fluid accumulation in the lungs creating a hazy or clouded area. \\
Emphysema & Enlarged air spaces in the lungs appearing over-expanded or damaged. \\
Enlarged Pulmonary Artery & Widening of the pulmonary artery seen as an enlarged artery in the chest. \\
Interstitial Lung Disease (ILD) & Scarring or inflammation of the lung’s interstitial tissue creating a reticular or nodular pattern. \\
Infiltration & Accumulation of substances or cells in the lung tissue visible as increased density or nodules. \\
Lung Cavity & Air-filled spaces within the lung often surrounded by dense tissue. \\
Lung Cyst & Fluid-filled spaces in the lung often round with thin walls. \\
Lung Opacity & An area of increased density in the lung fields typically appearing as a white or grayish patch. \\
Mediastinal Shift & Displacement of central chest structures like the heart to one side. \\
Nodule / Mass & A growth or lump in the lung which may appear as a well-defined or irregular shape. \\
Pulmonary Fibrosis & Scarring of the lung tissue creating a dense fibrous appearance. \\
Pneumothorax & Air trapped in the pleural space creating a gap or absence of lung tissue. \\
Pleural Thickening & Increased thickness of the pleura seen as a dense layer around the lung. \\
Pleural Effusion & Excess fluid in the pleural space appearing as a shadow around the lungs. \\
Rib Fracture & A break in one or more ribs appearing as a visible crack or displacement. \\
Other Lesion & An unusual mass or area in the lung with irregular borders or density. \\
\bottomrule
\end{tabular}
\end{table*}

\newpage
\begin{table*}[htbp]
\centering
\caption{Final attribute-based visual descriptions for each abnormality in PadChest-GR.}
\label{tab:attribute_padchest}
\renewcommand{\arraystretch}{1.3}
\begin{tabular}{p{5cm} p{10.5cm}}
\toprule
\textbf{Abnormality} & \textbf{Attribute-based Visual Description} \\
\midrule
Pleural Thickening & Increased thickness of the pleura seen as a dense layer around the lung. \\
Atelectasis & Collapsed lung tissue causing darkened or shrunken areas in the lung. \\
Pleural Effusion & Excess fluid in the pleural space appearing as a shadow around the lungs. \\
Cardiomegaly & Enlargement of the heart seen when the heart appears larger than normal. \\
Aortic Elongation & Lengthened and tortuous aorta, visible as an elongated curving structure. \\
Vertebral Degenerative Changes & Irregular vertebral margins with bony sclerosis and osteophytes. \\
Aortic Atheromatosis & Calcified deposits in the aortic wall appearing as bright, irregular opacities. \\
Nodule & A growth or lump in the lung which may appear as a well-defined or irregular shape. \\
Alveolar Pattern & Cloud-like, patchy opacities representing fluid or cellular accumulation in alveoli. \\
Hiatal Hernia & A soft-tissue mass or air-fluid level above the diaphragm, near the midline. \\
Scoliosis & Sideways curvature of the spine causing misalignment of vertebral bodies. \\
Hemidiaphragm Elevation & One side of the diaphragm appearing higher than the other, with convex shape. \\
Hyperinflated Lung & Abnormally increased lung volume with expanded air spaces. \\
Interstitial Pattern & Fine reticular or nodular opacities spread across the lung, indicating interstitial involvement. \\
Fracture & A break in the bone appearing as a radiolucent line or displacement. \\
Vascular Hilar Enlargement & Increased prominence of the pulmonary vessels near the lung hila. \\
NSG Tube & A thin radiopaque tube extending from the nasal cavity into the stomach. \\
Endotracheal Tube & A thin or opaque line in the middle of the trachea. \\
Hypoexpansion & Reduced lung inflation with increased density and narrow intercostal spaces. \\
Central Venous Catheter & A visible line inside large vein. \\
Electrical Device & A dense, well-defined metallic opacity, typically a pacemaker or defibrillator. \\
Bronchiectasis & Dilated bronchi with thick walls, appearing as tubular or cystic opacities. \\
Goiter & A soft tissue mass in the anterior neck, sometimes displacing the trachea. \\
Other lesions & An unusual mass or area in the lung with irregular borders or density. \\
\bottomrule
\end{tabular}
\label{tab:padchest_definitions}
\end{table*}

\newpage

\newpage

\end{document}